\documentclass[letterpaper]{article} 
\usepackage{aaai23}  
\usepackage{times}  
\usepackage{helvet}  
\usepackage{courier}  
\usepackage[hyphens]{url}  
\usepackage{graphicx} 
\usepackage{natbib}  
\usepackage{caption} 
\title{AutoInit: Analytic Signal-Preserving Weight Initialization for Neural Networks}
\author{
    Garrett Bingham\textsuperscript{\rm 1, 2} and Risto Miikkulainen \textsuperscript{\rm 1, 2}
}
\affiliations{
    \textsuperscript{\rm 1}
        The University of Texas at Austin,
        Austin, TX 78712 \\
    \textsuperscript{\rm 2}
        Cognizant AI Labs,
        San Francisco, CA 94105 \\
    bingham@cs.utexas.edu, risto@cs.utexas.edu
}

\usepackage[utf8]{inputenc} 
\usepackage{url}            
\usepackage{booktabs}       
\usepackage{amsfonts}       
\usepackage{nicefrac}       
\usepackage{microtype}      
\usepackage[table,dvipsnames]{xcolor}         

\usepackage{adjustbox}
\usepackage[ruled,vlined]{algorithm2e}
\usepackage{amsmath}
\usepackage{amssymb}
\usepackage{amsthm}
\usepackage{bm}
\usepackage{graphicx}
\usepackage{makecell}
\usepackage{mathtools}
\usepackage{multirow}
\usepackage{nicefrac}
\usepackage{subfigure}
\usepackage{tablefootnote}
\usepackage{tikz}

\newcommand{\inn}{\mathrm{in}}
\newcommand{\out}{\mathrm{out}}
\newcommand{\E}{\mathrm{E}}
\newcommand{\Var}{\mathrm{Var}}

\newcommand*\diff{\mathop{}\!\mathrm{d}}

\theoremstyle{plain}

\theoremstyle{definition}

\theoremstyle{remark}

\begin{document}

\maketitle

\begin{abstract}
  Neural networks require careful weight initialization to prevent signals from exploding or vanishing.  Existing initialization schemes solve this problem in specific cases by assuming that the network has a certain activation function or topology.  It is difficult to derive such weight initialization strategies, and modern architectures therefore often use these same initialization schemes even though their assumptions do not hold. This paper introduces AutoInit, a weight initialization algorithm that automatically adapts to different neural network architectures.  By analytically tracking the mean and variance of signals as they propagate through the network, AutoInit appropriately scales the weights at each layer to avoid exploding or vanishing signals.  Experiments demonstrate that AutoInit improves performance of convolutional, residual, and transformer networks across a range of activation function, dropout, weight decay, learning rate, and normalizer settings, and does so more reliably than data-dependent initialization methods.  This flexibility allows AutoInit to initialize models for everything from small tabular tasks to large datasets such as ImageNet.  Such generality turns out particularly useful in neural architecture search and in activation function discovery.  In these settings, AutoInit initializes each candidate appropriately, making performance evaluations more accurate. AutoInit thus serves as an automatic configuration tool that makes design of new neural network architectures more robust. The AutoInit package provides a wrapper around TensorFlow models and is available at \url{https://github.com/cognizant-ai-labs/autoinit}.
\end{abstract}

\section{Introduction}
\label{sec:introduction}

Proper weight initialization is crucial to achieve high performance with deep networks.  A common motif in such networks is repeated layers or building blocks.  Thus, if a given layer amplifies or diminishes the forward or backward propagation of signals, repeated applications of that layer will result in exploding or vanishing signals, respectively \cite{hochreiter1991untersuchungen, hanin2018neural}.  This phenomenon makes optimization difficult, and can even exceed machine precision. The issue persists regardless of whether the weights are sampled to be uniform, normal, or orthogonal \cite{saxe2013exact, hu2020provable}.

While many initialization strategies have been proposed in the past, these strategies apply only to neural networks with specific activation functions, topologies, or layer types.  Thus, researchers designing new models or activation functions have two options.  The first option is to derive weight initialization strategies manually for every architecture considered, which is generally difficult and time consuming.  The second option is to use existing initialization strategies in new settings, where they may be incorrect and therefore misleading:  A candidate model may appear poor when it is the suboptimal initialization that makes training difficult.

To overcome this problem, this paper proposes AutoInit, an algorithm that automatically calculates analytic mean- and variance-preserving weight initialization for neural networks.  Since AutoInit is algorithmic, it relieves the researcher from a difficult but consequential step in model design.  It is no longer necessary to use existing weight initialization strategies in incorrect settings: AutoInit provides an appropriate default initialization automatically, resulting in better and more reliable performance.



\section{Related Work}
\label{sec:related_work_extended}

Weight initialization strategies attempt to solve the following problem: How should weights be initialized so that the neural network does not suffer from vanishing or exploding signals?  This section reviews previous research in neural network weight initialization, which has focused on stabilizing signals by accounting for specific components of neural networks such as the activation function, topology, layer types, and training data distribution.

\paragraph{Activation-Function-Dependent Initialization}
\label{sec:weight_init_for_afns}
As is common in the literature, \texttt{fan\_in} and \texttt{fan\_out} refer to the number of connections feeding into and out of a node, respectively.  \citet{lecun2012efficient} recommend sampling weights from a distribution with mean zero and standard deviation $\sqrt{\texttt{fan\_in}}$.  This initialization encourages propagated signals to have variance approximately one if used with an activation function symmetric about the origin, like $1.7159\tanh\left(\frac{2}{3}x\right)$ or $\tanh(x) + \alpha x$ for some small choice of $\alpha$. The standard sigmoid $f(x) = 1/(1+e^{-x})$ induces a mean shift and should not be used in this setting.

\citet{glorot2010understanding} proposed one initialization strategy to ensure unit variance in the forward-propagated signals and another to ensure unit variance for the backward-propagated gradients.  As a compromise between the two strategies, they initialized weights by sampling from $\mathcal{U}\left(-\frac{\sqrt{6}}{\sqrt{\texttt{fan\_in} + \texttt{fan\_out}}}, \frac{\sqrt{6}}{\sqrt{\texttt{fan\_in} + \texttt{fan\_out}}}\right)$.  They also avoided sigmoid, and instead chose symmetric functions with unit derivatives at 0, such as tanh or Softsign$(x) = x/(1+|x|)$.

\citet{he2015delving} introduced the PReLU activation function and a variance-preserving weight initialization to be used with it that samples weights from $\mathcal{N}(0, \sqrt{2/\texttt{fan\_in}})$.  Similarly, \citet{selu} introduced SELU, an activation function with self-normalizing properties.  These properties are only realized when SELU is used with the initialization scheme by \citet{lecun2012efficient}.

The above weight initialization strategies attempt to solve the same fundamental problem: How can weights be scaled so that repeated applications of the activation function do not result in vanishing or exploding signals?  While these approaches solve this problem in a few special cases, the issue is more general.  Manually deriving the correct scaling is intractable for complicated activation functions.  One approach for an arbitrary function $f$ is to sample Gaussian inputs $x$ and adjust the weights according to the empirical variance $\textrm{Var}(f(x))$ \cite{brock2021characterizing}.  This paper proposes an alternative and potentially more accurate approach: integration by adaptive quadrature.  The result is a weight initialization strategy that is compatible with any integrable activation function.  Indeed, previous activation-function-dependent initializations are special cases of the AutoInit algorithm.

\paragraph{Topology-Dependent Initialization}
The activation-function-dependent initializations discussed above were designed for neural networks composed of convolutional or dense layers.  After the introduction of residual networks \citep[ResNets;][]{he2016deep, he2016identity}, new weight initialization schemes had to be developed to account for the effect of shortcut connections and various types of residual branches.

\citet{taki2017deep} analyzed signal propagation in plain and batch-normalized ResNets.  They developed a new weight initialization to stabilize training, but did not consider modifications like using deeper residual blocks or reordering components like the activation function or batch normalization layers.  In contrast, AutoInit is topology-agnostic: It adapts to any of these changes automatically.

\citet{zhang2019fixup} introduced Fixup, an initialization method that rescales residual branches to stabilize training.  Fixup replaces batch normalization in standard and wide residual networks \cite{ioffe2015batch, he2016deep, he2016identity, zagoruyko2016wide} and replaces layer normalization \cite{ba2016layer} in transformer models \cite{vaswani2017attention}.  The disadvantages of this scheme are that it only applies to residual architectures, needs proper regularization to get optimal performance, and requires additional learnable scalars that slightly increase model size.

\citet{arpit2019initialize} proposed a new initialization scheme for weight-normalized networks \cite{salimans2016weight} that relies on carefully scaling weights, residual blocks, and stages in the network.  Like related approaches, this technique improves performance in specific cases, but imposes design constraints, like requiring ReLU activation functions and a specific Conv $\rightarrow$ ReLU $\rightarrow$ Conv block structure.

Just as tanh-inspired weight initialization does not stabilize training of ReLU networks, initialization schemes designed for non-residual networks fail with ResNets \cite{hanin2018start, bachlechner2020rezero, brock2021characterizing}.
This observation suggests that future classes of neural networks will again require developing new weight initializations.  Additionally, practitioners with models that do not fit neatly within the restricted settings of existing weight initialization research are left to derive their own initialization or use a suboptimal one.  For example, many initialization schemes assume that the activation function is ReLU \cite{he2015delving, taki2017deep, arpit2019initialize, zhang2019fixup, de2020batch}.  Indeed, ReLU is currently the most popular activation function \cite{nwankpa2018activation, apicella2021survey}, but it is not the best choice in every case.  ReLU prevents dynamical isometry \cite{saxe2013exact, pennington2017resurrecting}, weakens adversarial training \cite{xie2020smooth}, and results in poorer accuracy compared to other activation functions in certain tasks \cite{bingham2020discovering}.  A general weight initialization strategy that does not impose architectural constraints and achieves good performance in diverse settings is needed. AutoInit is designed to meet this challenge.

\paragraph{Layer-Dependent Initialization}
\citet{hendrycks2016adjusting} noted that dropout layers \cite{srivastava2014dropout} also affect the variance of forward-propagated signals in a network.  To stabilize training properly, it is necessary to take dropout layers and the specific dropout rate into account in weight initialization.  In fact, pooling, normalization, recurrent, padding, concatenation, and other layer types affect the signal variance in a similar way, but current initialization schemes do not take this effect into account.  AutoInit is designed to adapt to each of these layer types dynamically, and can be extended to include new layer types as they are introduced in the future.

\paragraph{Data-Dependent Initialization}
\citet{mishkin2015all} fed data samples through a network and normalized the output of each layer to have unit variance.  \citet{krahenbuhl2015data} adopted a similar approach, but opted to normalize along the channel dimension instead of across an entire layer.  Data-dependent weight initializations are most similar in spirit to AutoInit; they rely on empirical variance estimates derived from the data in order to be model-agnostic.  However, data-dependent weight initializations introduce a computational overhead \cite{mishkin2015all}, and are not applicable in settings where data is not available or its distribution may shift over time, such as online learning or reinforcement learning.  The quality of the initialization is also dependent on the number of the data samples chosen, and suffers when the network is very deep \cite{zhang2019fixup}.  AutoInit instead uses an analytic approach for greater efficiency and higher accuracy.

\paragraph{Summary}
Thus, previous techniques solved the initialization problem for networks with specific activation functions, topologies, and layer types. In contrast, AutoInit does not impose design constraints, depend on data samples, or incur a parameter overhead \cite{dauphin2019metainit, zhu2021gradinit} and is therefore a good starting point especially in new settings.

\section{Neural Network Signal Propagation}

AutoInit aims to stabilize signal propagation throughout an entire neural network. More precisely, consider a layer that shifts its input by $\alpha$ and scales the input by a factor of $\beta$.  Given an input signal with mean $\mu_\inn$ and variance $\nu_\inn$, after applying the layer, the output signal will have mean $\mu_\out = \alpha + \beta \mu_\inn$ and variance $\nu_\out = \beta^2 \nu_\inn$.  In a deep network in which the layer is applied $L$ times the effect is compounded and the signal at the final layer has mean and variance
\begin{equation}\small
    \mu_\out = \beta^L\mu_\inn + \alpha(\beta^L +\beta^{L-1} + \cdots + \beta + 1), \;\; \nu_\out = \beta^{2L}\nu_\inn.
\end{equation}
If $|\beta| > 1$, the network will suffer from a mean shift and exploding signals as it increases in depth:
\begin{equation}\small
    \label{eq:explode}
    \lim_{L \rightarrow \infty} \mu_\out = \infty, \quad
    \lim_{L \rightarrow \infty} \nu_\out = \infty.
\end{equation}
In the case that $|\beta| < 1$, the network will suffer from a mean shift and vanishing signals:
\begin{equation}\small
    \label{eq:vanish}
    \lim_{L \rightarrow \infty} \mu_\out = \alpha / (1 - \beta), \quad
    \lim_{L \rightarrow \infty} \nu_\out = 0.
\end{equation}
AutoInit calculates analytic mean- and variance-preserving weight initialization so that $\alpha=0$ and $\beta=1$, thus avoiding the issues of mean shift and exploding/vanishing signals.

\section{The AutoInit Framework}
\label{sec:algorithm}

AutoInit is a general framework that adapts to different layer types.  Its implementation is outlined in Algorithm \ref{alg:autoinit}. A given \texttt{layer} in a neural network receives as its input a tensor $x$ with mean $\mu_\inn$ and variance $\nu_\inn$.  After applying the \texttt{layer}, the output tensor has mean $\mu_\out = \E(\texttt{layer}(x))$ and variance $\nu_\out = \Var(\texttt{layer}(x))$.  The function $g_{\texttt{layer}}$ maps input mean and variance to output mean and variance when the \texttt{layer} is applied:
\begin{equation}\small
    \label{eq:g}
    g_\texttt{layer} : (\mu_\inn, \nu_\inn) \mapsto (\mu_\out, \nu_\out).
\end{equation}
Note that $g$ in Equation \ref{eq:g} depends on the type of $\texttt{layer}$; e.g.\ $g_{\texttt{Dropout}}$ and $g_{\texttt{ReLU}}$ are different functions.  For layers with trainable weights, the mean and variance mapping will depend on those weights.  For example, the function $g_{\texttt{Conv2D},\theta}$ maps input mean and variance to output mean and variance after the application of a \texttt{Conv2D} layer parameterized by weights $\theta$.  Deriving $g$ for all layers makes it possible to model signal propagation across an entire neural network.  Thus, if $\mu_\inn$ and $\nu_\inn$ are known, it is natural to calculate initial weights $\theta$ such that the layer output will have zero mean and unit variance.  

\SetKwProg{Fn}{def}{:}{\textbf{end}}
\SetKwFunction{initlayer}{initialize}%
\begin{algorithm}
\small
\caption{AutoInit}
\KwIn{Network with layers $L$, directed edges $E$}
$\texttt{output\_layers} = \{l \in L \mid (l, l') \notin E \:\forall\: l' \in L \}$\\
\For{$\textrm{\tt output\_layer in output\_layers}$}{
    \texttt{initialize}(\texttt{output\_layer})
}
~\\
\Fn{\initlayer{\rm \tt layer}}{
    $\texttt{layers\_in} = \{l \in L \mid (l, \texttt{layer}) \in E\}$ \newline
    $i = 1$\\
    \For{$\textrm{\tt layer\_in in layers\_in}$}{
        $\mu_{\inn_i}, \nu_{\inn_i} = \texttt{initialize}(\texttt{layer\_in})$\\
        $i = i + 1$
    }
    $\mu_\inn = (\mu_{\inn_1}, \mu_{\inn_2}, \ldots, \mu_{\inn_N})$\\
    $\nu_\inn = (\nu_{\inn_1}, \nu_{\inn_2}, \ldots, \nu_{\inn_N})$\\
    \eIf{\textrm{\tt layer} {\rm has weights} $\theta$}{
        initialize $\theta$ s.t. $g_{\texttt{layer}, \theta}(\mu_\inn, \nu_\inn) = (0,1)$\\
        $\mu_\out, \nu_\out = 0, 1$
    }{
        $\mu_\out, \nu_\out = g_{\texttt{layer}}(\mu_\inn, \nu_\inn)$
    }
    \Return $\mu_\out, \nu_\out$
}
\label{alg:autoinit}
\end{algorithm}

For example, for \texttt{Conv2D} layers, one possibility is
\begin{equation}
    \label{eq:init_example}
    \scriptsize
    \theta \sim \mathcal{N}\left(0, 1/\sqrt{\texttt{fan\_in}(\nu_\inn + \mu_\inn^2)}\right) \! \implies \! g_{\texttt{Conv2D}, \theta} (\mu_\inn, \nu_\inn) = (0, 1).
\end{equation}

The AutoInit framework includes mean and variance mapping functions $g$ for the majority of layers used in modern architectures.  Appendix \ref{sec:mean_variance_estimation} details how these functions and the corresponding initialization strategies (e.g.\ Equation \ref{eq:init_example}) are derived.  New layers can be included by deriving $g$ manually, or by approximating it through Monte Carlo simulation.  This approach ensures that reliable estimates for $\mu_\inn$ and $\nu_\inn$ are available at all layers in a network, which in turn allows for weight initialization that stabilizes the signals to have zero mean and unit variance, avoiding the issues of mean shift and exploding/vanishing signals (Equations \ref{eq:explode} and \ref{eq:vanish}).


The main advantage of AutoInit is that it is a general method.  Unlike prior work, which imposes design constraints, AutoInit adapts to different settings automatically in order to improve performance.  The next seven sections demonstrate this adaptability experimentally from several perspectives: different classes of models (convolutional, residual, transformer), hyperparameter settings (activation function, dropout rate, weight decay, learning rate, optimizer), model depths (nine layer CNN to 812 layer ResNet), image sizes (ten-class $28 \times 28$ grayscale to 1{,}000-class $160 \times 160$ RGB), and data modalities (vision, language, tabular, multi-task, transfer learning).  AutoInit also outperforms data-dependent initialization methods and stabilizes convolutional, residual, and transformer networks without normalization layers.  This generality is shown to be particularly useful in neural architecture search and activation function discovery, where thousands of new designs need to be evaluated robustly.  AutoInit produces specialized weight initialization strategies for each candidate, which allows for measuring their performance more accurately. As a result, better solutions are discovered.  The experiments thus show that AutoInit is an effective initialization algorithm for existing networks as well as a good starting point for networks that may be developed in the future.

\section{Hyperparameter Variation in CNNs}
\label{sec:convolutional}

\begin{figure*}
    \centering
    \includegraphics[width=\linewidth]{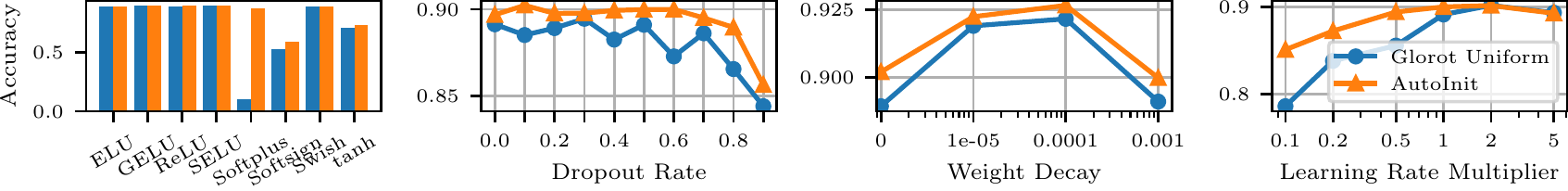}
    \caption{All-CNN-C test accuracy on CIFAR-10.  AutoInit results in comparable or better performance with different activation functions, better performance across all dropout rates and weight decay settings, and is less sensitive to the choice of learning rate than the default initialization.}
    \label{fig:allcnnc_hyperparams}
\end{figure*}

\begin{figure*}
    \centering
    \begin{tikzpicture}
    \draw (0, 0) node[anchor=south west, inner sep=0, align=left] {\includegraphics[width=\linewidth]{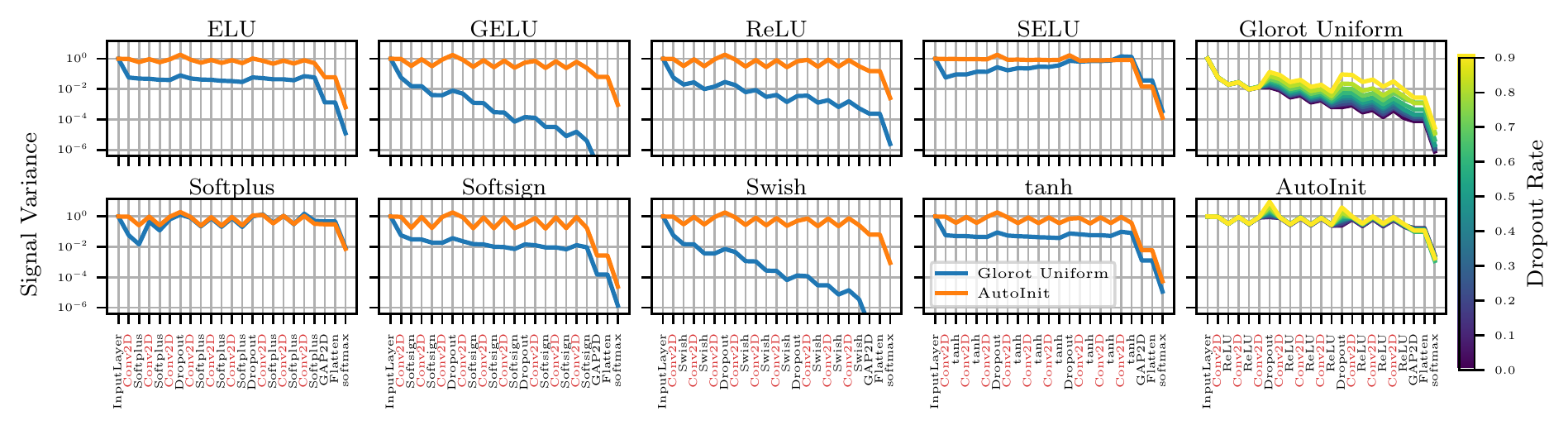}};
    \draw (0.6, 0) node[anchor=south west, inner sep=0, align=left] {(a)};
    \draw (10.1, 0) node[anchor=south west, inner sep=0, align=left] {(b)};
    \end{tikzpicture}
    \caption{Signal propagation in All-CNN-C networks with different (a) activation functions and (b) dropout rates.  With the default initialization, signals often vanish with depth, and their behavior is inconsistent across activation functions and dropout rates. With AutoInit, the variance fluctuates naturally as each layer modifies its input.  At layers with weights (marked in {\color{red}red}), AutoInit scales the weights appropriately to return the variance to approximately 1.0, stabilizing training in each case.}
    \label{fig:signal_propagation_allcnnc}
\end{figure*}

\paragraph{Experiment Setup}
The first experiment tests AutoInit's performance across a range of hyperparameter values for CNNs.  The experiment focuses on the All-CNN-C architecture \cite{springenberg2015striving}, which consists of convolutional layers, ReLU activation functions, dropout layers, and a global average pooling layer at the end of the network.  This simple design helps identify performance gains that can be attributed to proper weight initialization.  The network is trained on the CIFAR-10 dataset \cite{krizhevsky2009learning} using the standard setup (Appendix \ref{sec:training_details}).  In particular, the baseline comparison is the ``Glorot Uniform'' strategy \citep[also called Xavier initialization; ][]{glorot2010understanding}, where weights are sampled from $\mathcal{U}\left(-\frac{\sqrt{6}}{\sqrt{\texttt{fan\_in} + \texttt{fan\_out}}}, \frac{\sqrt{6}}{\sqrt{\texttt{fan\_in} + \texttt{fan\_out}}}\right)$.

\paragraph{Hyperparameter Variation}
In separate experiments, the activation function, dropout rate, weight decay, and learning rate multiplier were changed.  While one hyperparameter was varied, the others were fixed to the default values.

\paragraph{Results}
Figure \ref{fig:allcnnc_hyperparams} shows the performance of the network with the default initialization and with AutoInit in these different settings.  
In sum, AutoInit improved performance in every hyperparameter variation evaluated.  As Figure~\ref{fig:signal_propagation_allcnnc} shows, AutoInit is adaptive.  It alters the initialization to account for different activation functions and dropout rates automatically.

AutoInit is also robust.  Even as other hyperparameters like learning rate and weight decay change, AutoInit still results in a higher performing network than the default initialization.  The results thus suggest that AutoInit provides an improved default initialization for convolutional neural networks.

\section{Stability in Deep ResNets}
\label{sec:stability_deep_resnets}

This section expands the experimental analysis of AutoInit to residual networks, focusing on preactivation residual networks of various depths \cite{he2016identity}.  The training setup is standard unless explicitly stated otherwise (Appendix \ref{sec:training_details}).  In particular, the initialization is ``He Normal'' \cite{he2015delving}, where weights are sampled from $\mathcal{N}(0, \sqrt{2/\texttt{fan\_in}})$.  

\begin{figure}
    \centering
    \includegraphics[width=\linewidth]{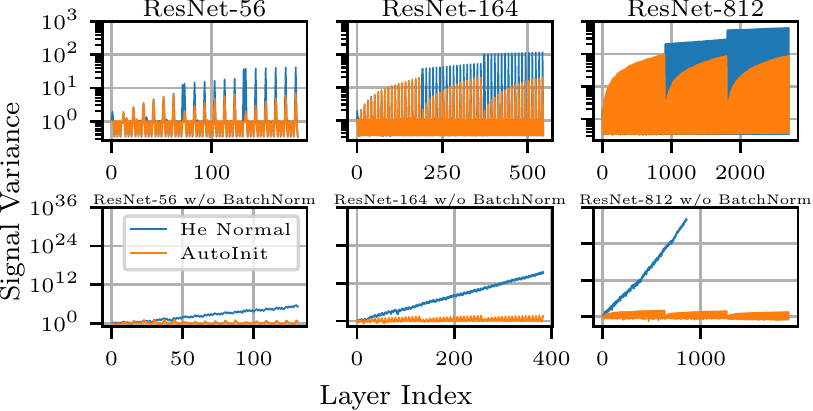}
    \caption{Signal propagation in residual networks.  Gaussian input was fed to the networks and empirical variance computed at each layer.  Since \texttt{ReLU}, \texttt{BatchNormalization}, and \texttt{Add} are counted as individual layers in this diagram, the total number of layers is different from that in the architecture name (i.e.\ ResNet-164 has 164 convolutional layers but over 500 total layers).  The default initialization causes exploding signals, while AutoInit ensures signal propagation is stable.}
    \label{fig:signal_propagation_resnet}
\end{figure}

\paragraph{Visualizing Signal Propagation} 
Figure \ref{fig:signal_propagation_resnet} shows how the signal variance changes with depth.  With ResNet-56, the variance increases where the shortcut connection and residual branch meet, and the variance drops whenever ReLU is applied.  Although the variance increases exponentially with the default initialization and linearly with AutoInit (note the log scale on the $y$ axis), training is still stable because batch normalization layers return the signal to variance 1.0.  Without batch normalization, the signal variance never stabilizes under the default initialization.  In contrast, removing batch normalization is not an issue with AutoInit; the signal variance remains stable with depth.

With the deeper ResNet-164 and ResNet-812 networks, the conclusions are similar but more pronounced.  In the case of ResNet-812 without batch normalization, the signals explode so severely that they exceed machine precision.  AutoInit avoids this issue entirely.

\paragraph{Stable Initial Learning Rates}
Exploding or vanishing signals make optimization difficult because they result in gradient updates that are too large to be accurate or too small to be meaningful.  This phenomenon can be observed when the network does not exceed chance accuracy. Therefore, a simple way to quantify whether a weight initialization is effective is to observe a network's performance after a few epochs.

\begin{figure}
    \centering
    \begin{tikzpicture}
    \draw (0, 0) node[anchor=south west, inner sep=0, align=left] {\includegraphics[width=\linewidth]{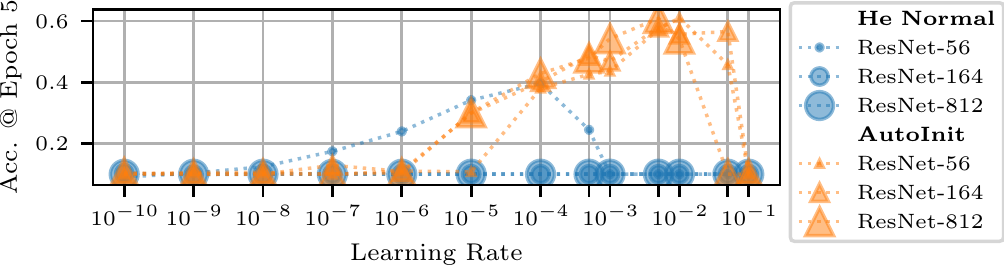}};
    \draw (0, 0) node[anchor=south west, inner sep=0, align=left] {(a)};
    \end{tikzpicture}\\[0.5em]
    
    \begin{tikzpicture}
    \draw (0, 0) node[anchor=south west, inner sep=0, align=left] {\includegraphics[width=\linewidth]{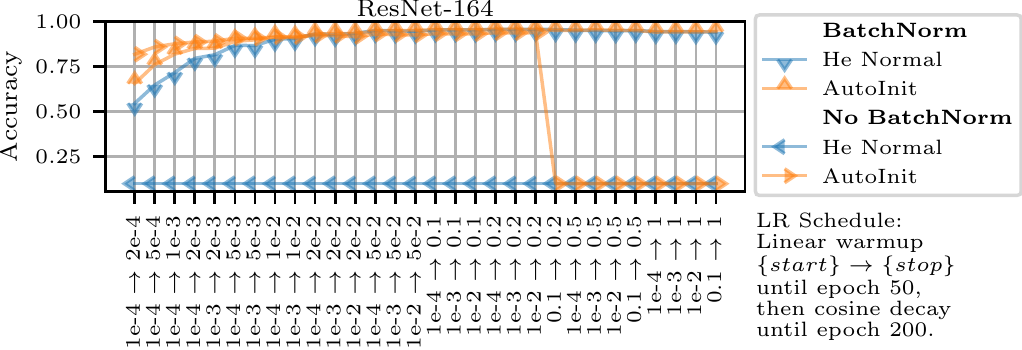}};
    \draw (0, 0) node[anchor=south west, inner sep=0, align=left] {(b)};
    \end{tikzpicture}

    \caption{ResNet accuracy on CIFAR-10 with different settings. (a) Accuracy of unnormalized ResNet architectures after five epochs of training with different learning rates and weight initializations.  While default initialization makes training difficult in ResNet-56 and impossible at greater depths, AutoInit results in consistent training at all depths.  (b) Accuracy of ResNet-164 with a variety of learning rate schedules and initializations. AutoInit is comparable to or outperforms the default initialization in every case.}
\label{fig:resnet_vary_initial_lr}
\end{figure}

Using this metric, AutoInit was compared against the default initialization by training unnormalized versions of ResNet-56, ResNet-164, and ResNet-812 for five epochs with a variety of learning rates.  With the default initialization, ResNet-56 requires a learning rate between $10^{-8}$ and $0.5 \times 10^{-3}$ to begin training, but training was not possible with ResNet-164 or ResNet-812 because of exploding signals (Figure \ref{fig:resnet_vary_initial_lr}a).  AutoInit stabilizes training for all three networks, and its effect does not diminish with depth.  The networks remain stable with higher learning rates between $10^{-4}$ and $0.05$.  Such rates speed up learning, and also correlate with better generalization performance \cite{jastrzkebski2017three, smith2018don, smith2018bayesian}.

\paragraph{Full ResNet Training}
In the third residual network experiment, ResNet-164 was trained to completion on CIFAR-10 with different learning rate schedules. All schedules included a linear warm-up phase followed by a decay to zero using cosine annealing \cite{loshchilov2016sgdr}.

Figure \ref{fig:resnet_vary_initial_lr}b displays the performance with a variety of such schedules.  When the best learning rate schedules are used, ResNet-164 achieves comparable performance with the default initialization and with AutoInit.  However, when a suboptimal schedule is used, performance degrades more quickly with the default initialization than it does with AutoInit.  Without batch normalization, the network requires proper weight initialization for stability.  In this case, ResNet-164 with the default initialization fails to train regardless of the learning rate schedule, whereas AutoInit results in high accuracy for the majority of them.

Together, the experiments in this section show that AutoInit is effective with deep networks.  It prevents signals from exploding or vanishing, makes it possible to use larger learning rates, and achieves high accuracy, with and without batch normalization. 

\section{High-Resolution Images with Transformers}
\label{sec:coatnet}
\begin{table}
    \centering
    \begin{adjustbox}{max width=\linewidth}
    \begin{tabular}{llllll}
        \toprule 
        CoAtNet & w/ GELU & w/ ReLU & w/ SELU & w/ Swish & w/o Norm \\ \midrule 
        Default Init.  & 89.38 & 89.22 & 86.09 & 88.69 & - \\
        Glorot Normal  & 91.44 & 91.54 & 87.59 & 90.42 & \textbf{85.89} \\
        Glorot Uniform & 91.16 & 91.18 & \textbf{88.25} & 90.06 & 85.73 \\
        He Normal      & 88.48 & 88.05 & 86.11 & 88.36 & - \\
        He Uniform     & 88.66 & 87.87 & 86.37 & 88.41 & - \\
        LeCun Normal   & 91.11 & 90.57 & 87.80 & 90.83 & - \\
        LeCun Uniform  & 90.55 & 90.65 & 87.67 & 90.57 & - \\
        AutoInit & \textbf{92.48} & \textbf{92.15} & 86.80 & \textbf{92.28} & 85.73 \\ 
        \bottomrule
    \end{tabular}
    \end{adjustbox}
    \caption{CoAtNet top-1 accuracy on Imagenette, shown as median of three runs.  The first four experiments vary the activation function, while the fifth removes all normalization layers from the architecture. A ``-'' indicates that training diverged.  AutoInit produces the best model in three of the five settings, and remains stable even without normalization layers.}
    \label{tab:coatnet}
\end{table}
This section extends AutoInit to transformer architectures and applies them to high-resolution image classification.  Specifically, AutoInit is applied to CoAtNet, a model that combines convolutional and attention layers \cite{dai2021coatnet}.  The model is trained on Imagenette, a subset of 10 classes from the ImageNet dataset \cite{imagenette, deng2009imagenet}.  Imagenette allows evaluating AutoInit in a high-resolution image classification task with a $132\times$ smaller carbon footprint than the full ImageNet dataset would (Appendix \ref{sec:compute_infrastructure}).  As Table \ref{tab:coatnet} shows, AutoInit outperforms six commonly used initialization schemes as well as the default initialization, which initializes convolutional layers from $\mathcal{N}(0,\sqrt{2/\texttt{fan\_out}})$ and fully-connected layers from $\mathcal{U}\left(-\frac{\sqrt{6}}{\sqrt{\texttt{fan\_in} + \texttt{fan\_out}}}, \frac{\sqrt{6}}{\sqrt{\texttt{fan\_in} + \texttt{fan\_out}}}\right)$.  Furthermore, AutoInit stabilizes the network even when normalization layers are removed, suggesting that it is a promising candidate towards developing normalizer-free transformer architectures.  Full experiment details are in Appendix \ref{sec:training_details}.

\section{Scaling up to ImageNet}
\begin{table}
    \centering
    \begin{adjustbox}{max width=\linewidth}
    \begin{tabular}{lll}
        \toprule 
        & top-1 & top-5 \\ \midrule 
        Default Init. & 74.33 & 91.60 \\
        AutoInit & \textbf{75.35} & \textbf{92.03} \\
        \bottomrule
    \end{tabular}
    \end{adjustbox}
    \caption{ResNet-50 top-1 and top-5 validation accuracy on ImageNet.  AutoInit improves performance, even with large and challenging datasets.}
    \label{tab:imagenet}
\end{table}

\label{sec:imagenet}

In order to compliment the results from Section \ref{sec:coatnet} and demonstrate that AutoInit can scale to more difficult tasks, ResNet-50 was trained from scratch on ImageNet with the default initialization and with AutoInit.  As Table \ref{tab:imagenet} shows, AutoInit improves top-1 and top-5 accuracy in this task as well.  Full training details are in Appendix \ref{sec:training_details}.

\section{\hspace{-1ex}Contrast with Data-Dependent Initialization}
\begin{figure}
    \centering
    \includegraphics[width=\linewidth]{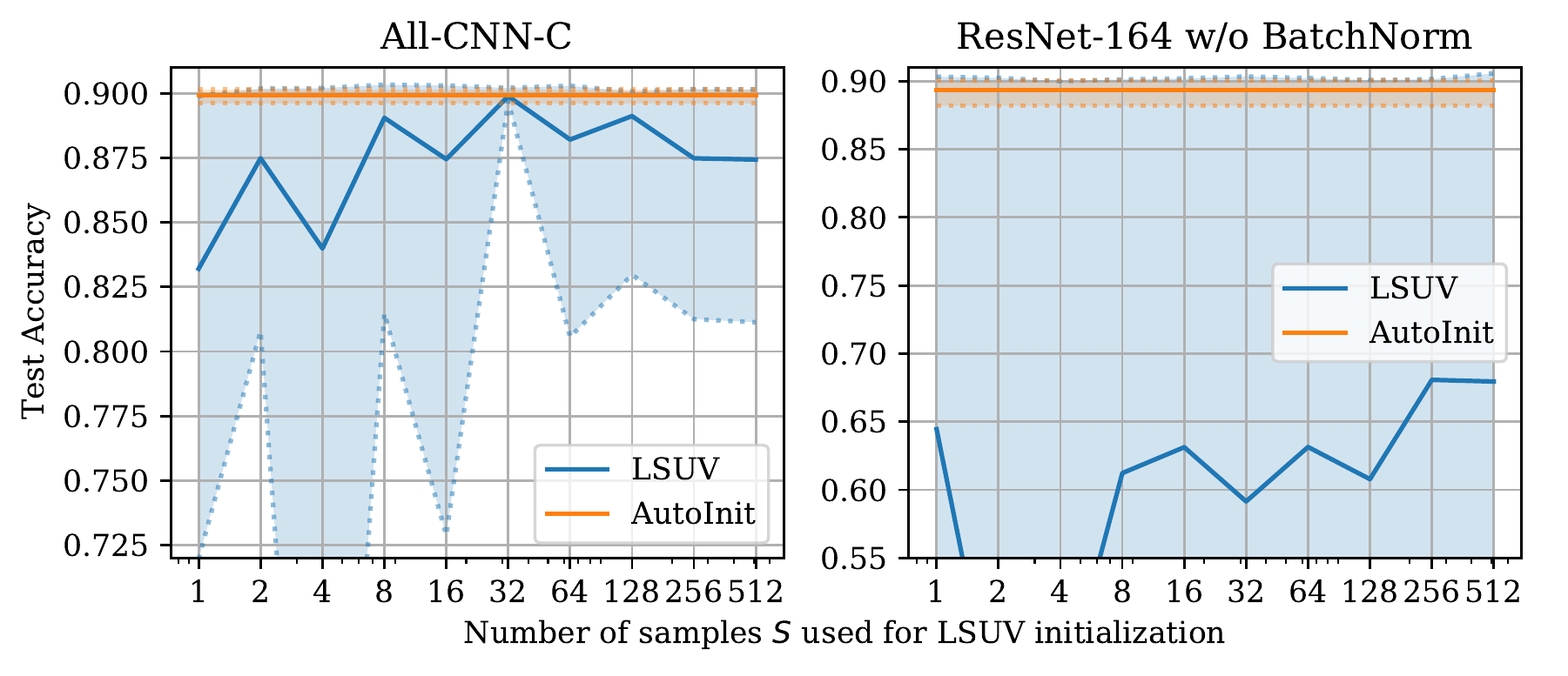}
    \caption{Mean CIFAR-10 test accuracy for AutoInit vs.\ LSUV with different numbers of samples $S$.  Each evaluation is repeated 10 times; the shaded area shows the maximum and minimum accuracy among all trials.  AutoInit is consistent, but LSUV struggles when $S$ is small or the network is deep.}
    \label{fig:performance}
\end{figure}

\label{sec:autoinit_vs_lsuv}
The layer-sequential unit-variance (LSUV) algorithm is the most natural data-dependent initialization comparison to AutoInit because both approaches aim to scale the weights appropriately in an architecture-agnostic way. LSUV pre-initializes the weights with an existing approach, feeds $S$ training samples through the network, and adjusts the scale of the weights so that each layer's output variance is approximately one \cite{mishkin2015all}.

Data-dependent initialization is time-consuming for large $S$ (indeed, even $S=1$ is used in practice \cite{kingma2018glow}).  However, if $S$ is too small, the samples may not reflect the statistics of the dataset accurately, leading to poor initialization.  Figure \ref{fig:performance} demonstrates this phenomenon.  In some training runs LSUV matches the performance of AutoInit, but in many instances the randomly selected samples do not accurately reflect the overall dataset and performance suffers.  Since AutoInit is not data-dependent, it does not have this issue.  Details of this experiment are in Appendix \ref{sec:training_details}.

\section{Enabling Neural Architecture Search}
\label{sec:nas}

Sections \ref{sec:convolutional} through \ref{sec:autoinit_vs_lsuv} demonstrated that AutoInit works well for convolutional, residual, and transformer networks with a variety of hyperparameter values and depths. In this section, the results are extended to a broader variety of network topologies and types of tasks, for two reasons. First, whereas custom weight initialization may be developed by hand for the most popular machine learning benchmarks, it is unlikely to happen for a variety of architectures and tasks beyond them. Second, as new types of neural network designs are developed in the future, it will be important to initialize them properly to reduce uncertainty in their performance.  This section evaluates the generality of AutoInit by applying it to the variety of networks generated in a neural architecture search process with five types of tasks.

\usetikzlibrary{shapes.geometric, arrows, positioning, calc}
\tikzstyle{input} = [rectangle, rounded corners, text centered, draw=black, fill=red!30]
\tikzstyle{unary} = [rectangle, rounded corners, text centered, draw=black, fill=yellow!30]
\tikzstyle{binary} = [rectangle, rounded corners, text centered, draw=black, fill=blue!30]
\tikzstyle{arrow} = [thick, ->,>=stealth]
\tikzstyle{output} = [rectangle, rounded corners, text centered, draw=black, fill=green!30]
\tikzstyle{invisible} = [rectangle, text centered]
\tikzstyle{maybeparam} = [draw, circle, dashed, fill=cyan!30]

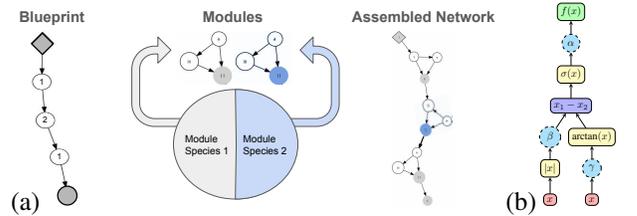
\begin{figure}[!t]
    \centering
    \begin{tikzpicture}
    \draw (0, 0) node[anchor=south west, inner sep=0, align=left] {\includegraphics[width=0.77\linewidth]{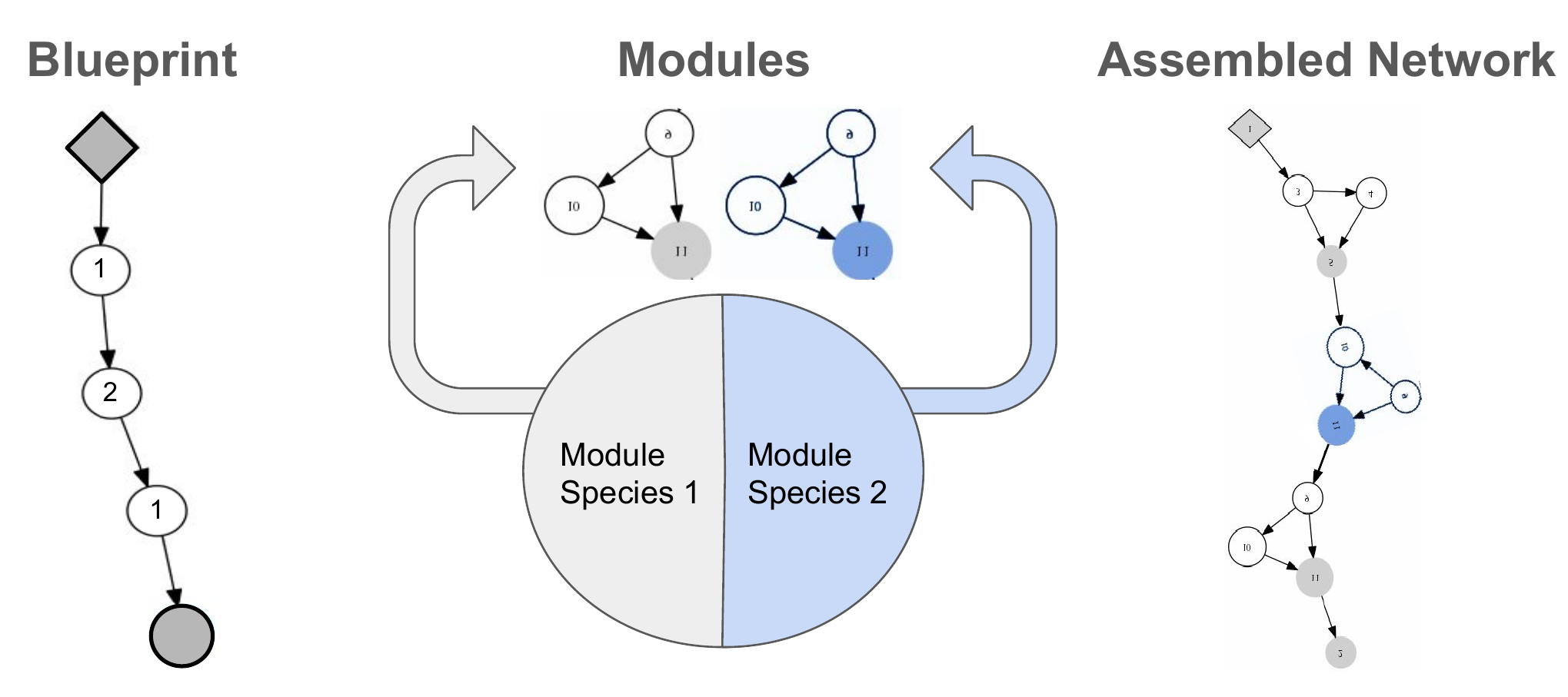}};
    \draw (0, 0) node[anchor=south west, inner sep=0, align=left] {(a)};
    \end{tikzpicture}
    \begin{tikzpicture}
    \draw (0, 0) node[anchor=south west, align=left] {
    
        \begin{adjustbox}{max width=0.12\linewidth}
            \begin{tikzpicture}[node distance=3em]
        
                    \node (output) [output] {$f(x)$};
                    \node (p1) [maybeparam, below of=output] {$\alpha$};
                    \node (unary) [unary, below of=p1] {$\sigma(x)$};
                    \node (binary) [binary, below of=unary] {$x_1 - x_2$};
                    \node (p5) [maybeparam, below of=binary, xshift=-2em] {$\beta$};
                    \node (unary1) [unary, below of=p5] {$|x|$};
                    \node (unary2) [unary, below of=binary, xshift=2em] {$\textrm{arctan}(x)$};
                    \node (p4) [maybeparam, below of=unary2] {$\gamma$};
                    \node (input1) [input, below of=unary1] {$x$};
                    \node (input2) [input, below of=p4] {$x$};

                    \draw [arrow] (input1) -- (unary1);
                    \draw [arrow] (input2) -- (p4);
                    \draw [arrow] (p4) -- (unary2); 
                    \draw [arrow] (unary1) -- (p5);
                    \draw [arrow] (p5) -- (binary);
                    \draw [arrow] (unary2) -- (binary);
                    \draw [arrow] (binary) -- (unary);
                    \draw [arrow] (unary) -- (p1);
                    \draw [arrow] (p1) -- (output);
                    
                \end{tikzpicture}
        \end{adjustbox}
        
    };
    \draw (-0.33, 0) node[anchor=south west, inner sep=0, align=left] {(b)};
    \end{tikzpicture}

    \caption{(a) The CoDeepNEAT method.  Modules replace nodes in the blueprint to create a candidate neural network. (b) An example activation function created with the PANGAEA method.  The computation graph represents the parametric function $f(x) = \alpha \cdot \sigma (\beta \cdot | x | - \arctan ( \gamma \cdot x ) )$.  CoDeepNEAT and PANGAEA generate a variety of architectures and activation functions that can be used to evaluate AutoInit's generality and flexibility.}
    \label{fig:codeepneat_pangaea}
\end{figure}

\begin{figure*}
    \centering

    \begin{tikzpicture}
    \draw (0, 0) node[anchor=south west, inner sep=0, align=left] {\includegraphics[width=\linewidth]{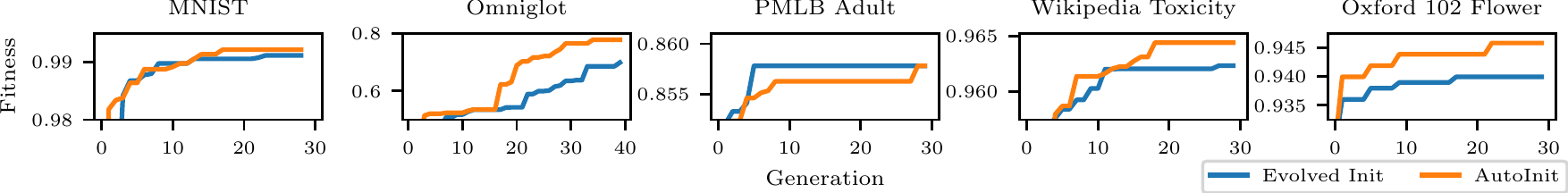}};
    \draw (0, 0) node[anchor=south west, inner sep=0, align=left] {(a)};
    \end{tikzpicture}\\
    
        \begin{tikzpicture}
            \draw (0, 0) node[anchor=south west, inner sep=0, align=left] {\includegraphics[width=182pt]{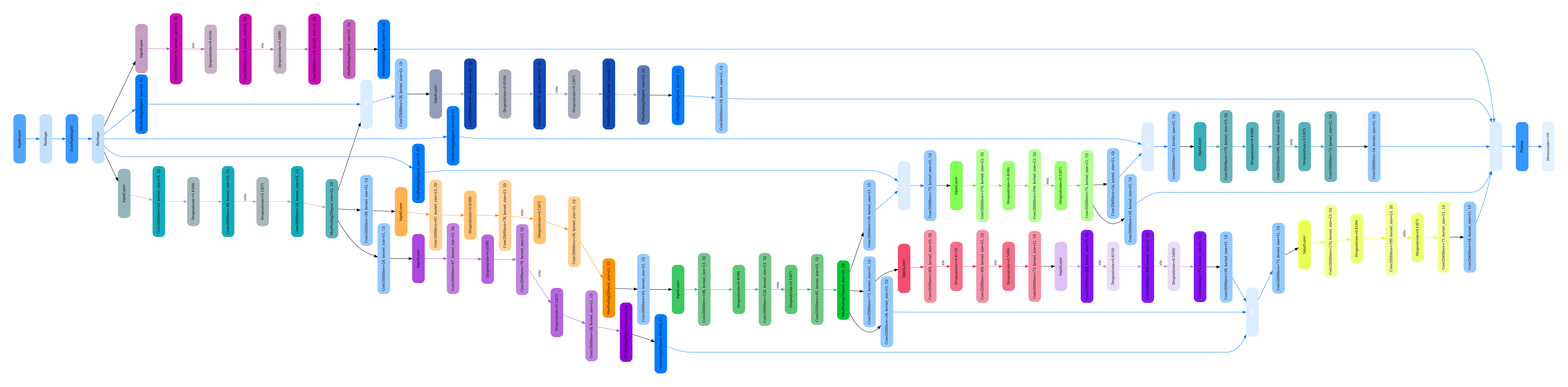}
            \includegraphics[width=104pt]{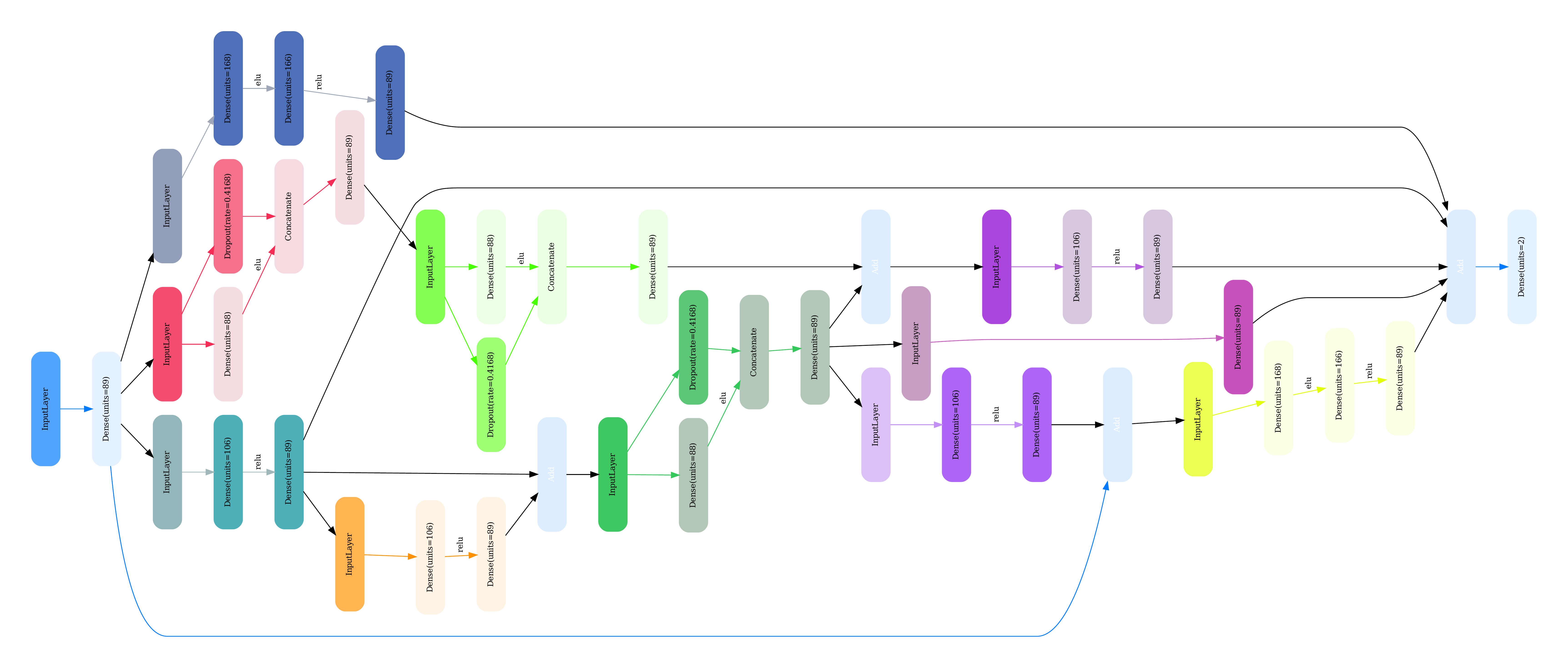}
            \includegraphics[width=212pt]{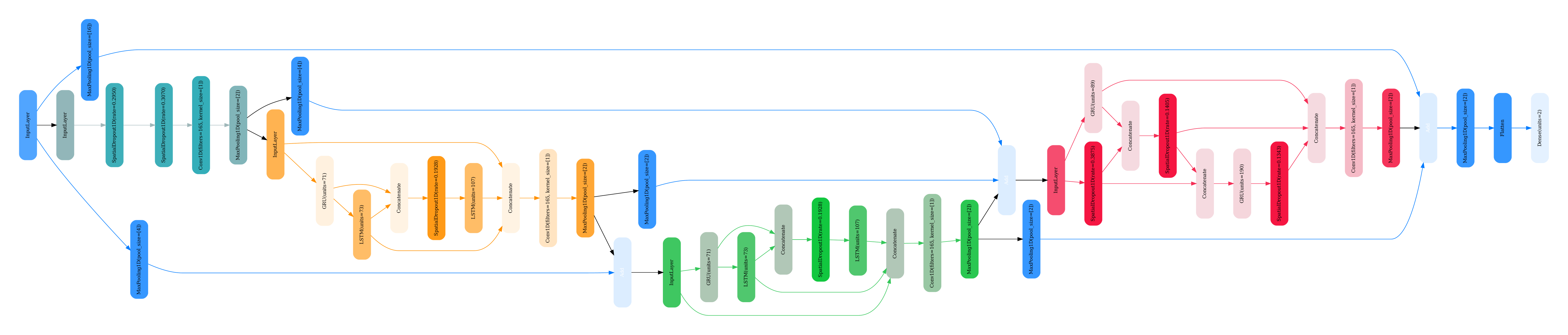}\\
            \adjustbox{max width=\linewidth}{
                \begin{tabular}{p{182pt}<{\centering}p{104pt}<{\centering}p{212pt}<{\centering}}
                \textbf{MNIST} & \textbf{PMLB Adult} & \textbf{Wikipedia Toxicity}
                \end{tabular}}
            };
        \draw (0, 0) node[anchor=south west, inner sep=0, align=left] {(b)};
        \end{tikzpicture}

    
    \caption{Evaluation of AutoInit with neural architecture search. (a) Performance improvement over generations in the five tasks.  AutoInit outperforms the evolved initialization on four tasks and matches it on one.  (b) Representative networks evolved with AutoInit.  Although the networks are distinct, AutoInit initializes them properly, leading to good performance in each case.
    }
    \label{fig:enn_results}
\end{figure*}

\paragraph{The CoDeepNEAT Architecture Search Method}
Neural networks are evolved using CoDeepNEAT \cite{liang2019evolutionary, miikkulainen2019evolving}.  
CoDeepNEAT extends previous work on evolving network topologies and weights \cite{moriarty1997forming, stanley2002evolving} to the level of evolving deep learning architectures.  Utilizing a cooperative coevolution framework \cite{potter2000cooperative}, CoDeepNEAT evolves populations of modules and blueprints simultaneously (Figure \ref{fig:codeepneat_pangaea}a).  Modules are small neural networks, complete with layers, connections, and hyperparameters.  Blueprints are computation graphs containing only nodes and directed edges.  To create a candidate neural network, CoDeepNEAT chooses a blueprint and replaces its nodes with selected modules.  This mechanism makes it possible to evolve deep, complex, and recurrent structures, while taking advantage of the modularity often found in state-of-the-art models.  In addition to the network structure, CoDeepNEAT evolves hyperparameters like dropout rate, kernel regularization, and learning rate.  The network weights are not evolved, but instead trained with gradient descent.  The generality of CoDeepNEAT helps minimize human design biases and makes it well-suited to analyzing AutoInit's performance in a variety of open-ended machine learning settings.

\paragraph{Tasks}
Using CoDeepNEAT, networks are evolved for their performance in vision (MNIST), language (Wikipedia Toxicity), tabular (PMLB Adult), multi-task (Omniglot), and transfer learning (Oxford 102 Flower) tasks (Appendix~\ref{sec:enn_details}).

\paragraph{Results}
Figure \ref{fig:enn_results}a shows how CoDeepNEAT discovers progressively better networks over time on the five tasks.  Evolution often selects different weight initialization strategies for the different layers in these networks, so this scheme is already a flexible and powerful baseline.  However, by accounting for each model's unique topology and hyperparameters, AutoInit outperforms the baseline in four of the five tasks, and matches it in the fifth.

Beyond performance, three interesting phenomena can be observed. First, the mean population fitness varies greatly with the default initialization in each task, sometimes dropping significantly from one generation to the next (Figure \ref{fig:enn_results_detail} in Appendix \ref{sec:enn_details}).  Though some variation is natural in a stochastic evolutionary process like CoDeepNEAT, AutoInit makes the discovery process more reliable by stabilizing the performance of the entire population.

Second, hyperparameters play a large role in the final performance of the dense networks, in particular in the ``Oxford 102 Flower'' task.  While CoDeepNEAT discovers good models with both initialization strategies, performance is consistently higher with AutoInit. This finding agrees with Section \ref{sec:convolutional}, where AutoInit was shown to be robust to different hyperparameter values.  

Third, while many networks exhibit motifs popular in existing architectures, such as alternating convolution and dropout layers and utilizing residual connections, other phenomena are less common (Figure \ref{fig:enn_results}b).  For example, the networks make use of different activation functions and contain several unique information processing paths from the input to the output.  Because AutoInit provides effective initialization in each of these cases, it allows for taking full advantage of unusual design choices that might otherwise hurt performance under default initialization schemes.

The results in this section suggest that AutoInit is an effective, general-purpose algorithm that provides a more effective initialization than existing approaches when searching for new models.

\section{Enabling~Activation~Function~Discovery}
\label{sec:afn_meta_learning}

\begin{figure}[ht]
    \centering
    \begin{tikzpicture}
    \draw (0, 0) node[anchor=south west, inner sep=0, align=left] {\includegraphics[width=\linewidth]{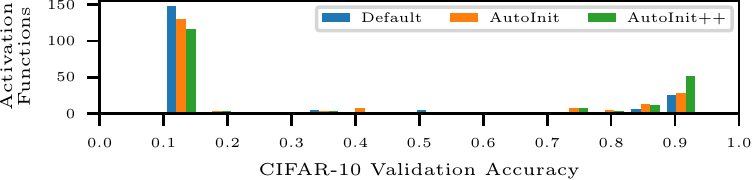}};
    \draw (0, 0) node[anchor=south west, inner sep=0, align=left] {(a)};
    \end{tikzpicture}\\[1em]
    \begin{tikzpicture}
    \draw (0, 0) node[anchor=south west, inner sep=0, align=left] {\includegraphics[width=\linewidth]{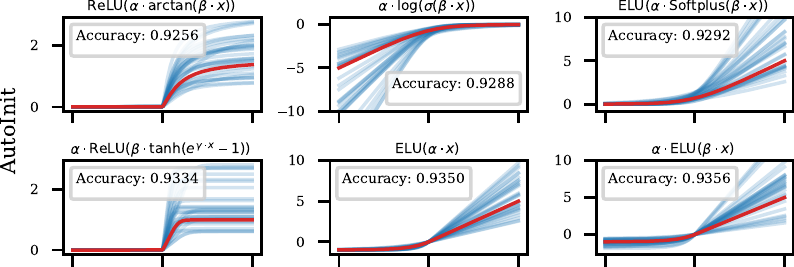}\\[1em]
    \includegraphics[width=\linewidth]{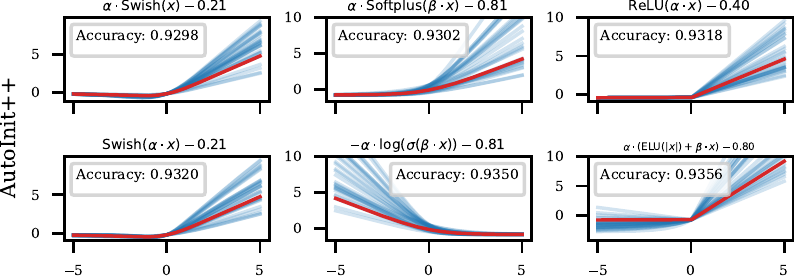}};
    \draw (0, 0) node[anchor=south west, inner sep=0, align=left] {(b)};
    \end{tikzpicture}
    
    \caption{Evaluation of AutoInit with activation function discovery. (a) Distribution of accuracies achieved with 200 activation functions and different weight initialization strategies. AutoInit and AutoInit++ make training more stable and allow more high-performing activation functions to be discovered than the default initialization does.  (b) High-performing activation functions.  The red line shows the function at initialization, with $\alpha = \beta = \gamma = 1$.  The blue lines show the shapes the activation function takes during training, created by sampling $\alpha, \beta, \gamma$ from $\mathcal{U}(0.5, 2.0)$.  AutoInit's flexibility should turn out useful for developing new activation functions in the future.}
    \label{fig:random_afns_hist}
\end{figure}
As new activation functions are developed in the future, it will be important to adjust weight initialization to maintain stable signal propagation.  Since AutoInit makes this adjustment automatically, it is well-suited to the task.  Indeed, Figure \ref{fig:allcnnc_hyperparams} confirmed that AutoInit improves performance with several existing activation functions.  This section presents a more challenging task.  To simulate future research in activation function design, hundreds of novel activation functions were generated as arbitrary computation graphs and trained with a CNN.  AutoInit's ability to initialize each of these networks was then evaluated.  The method for creating such activation functions is described first, followed by experimental details, and results on stability, performance, and generality.

\paragraph{Creating Novel Activation Functions}
An important area of automated machine learning (AutoML) is to discover new, better activation functions \cite{basirat2018quest, DBLP:conf/iclr/RamachandranZL18, bingham2020gecco, liu2020evolving}.
Among existing approaches, PANGAEA \cite{bingham2020discovering} has the most flexible search space and is therefore used to generate new functions in this section.

PANGAEA represents activation functions as computation graphs containing unary and binary operators (Figure \ref{fig:codeepneat_pangaea}b).  Creating a novel activation function involves three steps.  First, a minimal computation graph is initialized with randomly selected unary and binary operators.  Second, the functional form of the activation function is modified by applying three random mutations to increase diversity.  Third, the function is augmented with up to three learnable parameters.  These parameters are analogous to those in other parametric activation functions, such as PReLU \cite{he2015delving}; they are initialized to one and learned during training by gradient descent.  Through this process, it is possible to understand to what extent AutoInit can improve performance with activation functions that have yet to be discovered.

\paragraph{Experimental Setup}
An important insight in this domain is that in addition to modifying the variance of the signals in a network, activation functions can induce mean shifts.  Prior work encouraged stability by reparameterizing the weights to have zero empirical mean \cite{huang2017centered, qiao2019micro, brock2021characterizing}.  An alternative and more direct approach is to modify the activation function itself so that it does not cause a mean shift in the first place.  Given an activation function $f$ with Gaussian mean $\mu_f = \frac{1}{\sqrt{2\pi}}\int_{-\infty}^\infty f(x) e^{-x^2/2} \diff x$, this goal can be accomplished with $\tilde{f} \coloneqq f - \mu_f$, which has zero Gaussian mean. To take advantage of this idea, a version of AutoInit called AutoInit++ was created for this domain, thus extending AutoInit slightly beyond weight initialization.

Thus, three initialization strategies were compared.  With the default initialization, weights were sampled from $\mathcal{U}\left(-\frac{\sqrt{6}}{\sqrt{\texttt{fan\_in} + \texttt{fan\_out}}}, \frac{\sqrt{6}}{\sqrt{\texttt{fan\_in} + \texttt{fan\_out}}}\right)$ \cite{glorot2010understanding}.  With AutoInit, the weights were sampled from $\mathcal{N}\left(0, 1/\sqrt{\texttt{fan\_in}\mu_f}\right)$ to account for an arbitrary activation function $f$; the dropout adjustment (Section~\ref{sec:convolutional}) was not used. Finally, AutoInit++ takes advantage of $\tilde{f}$ as described above, but is otherwise identical to AutoInit.

For each initialization strategy, 200 activation functions were created using the PANGAEA process.  Each activation function was used with the All-CNN-C architecture on the CIFAR-10 dataset following the standard training setup.  To avoid overfitting to the test set when evaluating such a large number of activation functions, the accuracy with a balanced validation set of 5000 images is reported instead.

\paragraph{Stability}
Achieving better-than-chance accuracy is a useful metric of training stability (Section \ref{sec:stability_deep_resnets}).  As shown in Figure \ref{fig:random_afns_hist}a, many activation functions result in chance accuracy regardless of how the network is initialized.  This phenomenon is not surprising; since the activation functions are arbitrary computation graphs, many of them will turn out to be poor.  With the default initialization strategy, 149 activation functions caused training to fail in this way.  With AutoInit, the number of failed activation functions dropped to 130, and with AutoInit++, it further decreased to 117.  AutoInit and AutoInit++ thus make training more stable, allowing it to succeed for a greater number of activation functions.

\paragraph{Performance}
Beyond training stability, a good weight initialization should also improve performance.  As a baseline, when trained with ReLU and the default initialization, All-CNN-C achieved 89.10\% test accuracy.  Twenty-two of 200 activation functions from the PANGAEA search space outperformed this accuracy with the default initialization.  With AutoInit, this number increased to 26, and with AutoInit++, to 50---a notable improvement.  Thus, with the default initialization, one can naively create a randomly generated computation graph activation function and have roughly a one in nine chance of outperforming ReLU, but with AutoInit++, this probability increases to one in four.

Indeed, the Mann-Whitney U test \cite{mann1947test} concludes that the distribution of accuracies induced by AutoInit++ is \textit{stochastically larger} than that from AutoInit $(p < 0.05)$ or the default initialization $(p < 0.01)$.  This result means that for any level of performance, it is always more probable to discover an activation function that achieves that level of performance when initializing with AutoInit++ versus AutoInit or the default initialization.  The result implies that activation function researchers who properly initialize their networks are more likely to discover state-of-the-art activation functions, while staying with the default initialization may hinder that research effort.  More detailed statistical significance analyses are included in Appendix \ref{sec:stat_sig}.

\paragraph{Generality}
Figure \ref{fig:random_afns_hist}b plots several activation functions from the PANGAEA search space.  Many discovered functions have similar shapes to existing functions.  However, others are nonmonotonic, have discontinuous derivatives, or saturate to nonzero values.  These properties are less common in existing activation functions.  This observation suggests that AutoInit is a general approach that does not depend on a specific type of activation function; it may therefore serve as a useful tool in developing new such functions in the future.

\section{Future Work}
\label{sec:future_work}

\paragraph{Experiments in Other Domains}
The experiments in this paper demonstrate that AutoInit can improve performance in a variety of settings, suggesting that it can be applied to other domains as well.  For instance in reinforcement learning, good estimates of activation statistics are usually not available due to the online nature of the algorithm.  It is not possible to stabilize training using e.g.\ batch normalization, but it may be possible to do it with AutoInit.  Similarly, training of generative adversarial networks \cite{goodfellow2014generative} is often unstable, and proper initialization may help.  
Applying AutoInit to such different domains should not only make them more reliable, but also lead to a better understanding of their training dynamics.

\paragraph{Accelerating Model Search}
In Sections~\ref{sec:nas} and~\ref{sec:afn_meta_learning},  AutoInit was shown to facilitate the discovery of better neural network designs and activation functions. This ability is possible because AutoInit is a general method, i.e.\ not restricted to a single class of models, and it could similarly augment other meta-learning algorithms \citep[e.g.\ those reviewed by][]{elsken2019neural, wistuba2019survey}.

However, this finding points to an even more promising idea.  As model search techniques become more prevalent in real-world applications, it will be most worthwhile to derive general principles rather than specific instantiations of those principles.  For example, past weight initialization strategies improved performance with specific activation functions through manual derivation of appropriate weight scaling (Section~\ref{sec:weight_init_for_afns}). In contrast, AutoInit is a general method, leveraging Gaussian quadrature for any activation function.  Similarly, AutoInit resulted in better initialization than strategies discovered by CoDeepNEAT through evolution (Section~\ref{sec:nas}).  Further, AutoInit++ (Section~\ref{sec:afn_meta_learning}), rather than producing a few high-performing activation functions, introduces the general property that activation functions with zero Gaussian mean ($\tilde{f} \coloneqq f - \mu_f$) tend to perform well.  This property discovered a highly diverse set of powerful activation functions in the PANGAEA search space (Figure \ref{fig:random_afns_hist}).  

Thus, AutoInit is successful because it is not a single initialization strategy, but rather a mapping from architectures to initialization strategies.  Such mappings, whether focused on initialization or some other aspect of model design, deserve increased attention in the future.  They can lead to performance gains in a variety of scenarios.  They also accelerate model search by focusing the search space to more promising regions.  If one does not have to worry about discovering a good initialization, compute power can instead be used in other areas, like designing architectures and activation functions.  Thus, general tools like AutoInit save time and resources, and lead to better models as a result.

Further technical extensions to AutoInit are outlined in Appendix~\ref{sec:technical_extensions}), including variations on the core AutoInit algorithm, support for new layer types, and integration with deep learning frameworks.
\section{Conclusion}
\label{sec:conclusion}

This paper introduced AutoInit, an algorithm that calculates analytic mean- and variance-preserving weight initialization for neural networks automatically.  In convolutional networks, the initialization improved performance with different activation functions, dropout rates, learning rates, and weight decay settings.  In residual networks, AutoInit prevented exploding signals, allowed training with higher learning rates, and improved performance with or without batch normalization.  In transformers, AutoInit was scaled up to high-resolution image classification, and improved performance with several activation functions with and without normalization.  AutoInit also improved accuracy on the ImageNet dataset.  The initialization is independent of data and is therefore efficient and reliable. AutoInit's generality proved instrumental in two types of AutoML. In neural architecture search, new architectures were evaluated more accurately, resulting in better networks in vision, language, tabular, multi-task, and transfer learning settings.  In activation function discovery, AutoInit stabilized training and improved accuracy with a large diversity of novel activation functions.  Thus, AutoInit serves to make machine learning experiments more robust and reliable, resulting in higher performance, and facilitating future research in AutoML.

\bibliography{references}
\newpage
\appendix

\section{Mean and Variance Estimation for Different Layer Types}
\label{sec:mean_variance_estimation}

In the AutoInit framework of Algorithm~\ref{alg:autoinit}, the mean and variance mapping function $g$ needs to be defined for each type of layer in a given neural network.  This appendix presents derivations for a majority of the most commonly used layers available in the TensorFlow package \cite{abadi2016tensorflow} at time of writing.  Extending AutoInit to support new layers in the future will require deriving the function $g$ for those layers.  Monte Carlo sampling can also be used as an approximation for $g$ before it is manually derived.  

In the following paragraphs, $x$ denotes the input to a layer, and $y$ is the output.  The incoming and outgoing means and variances are denoted as $\mu_\inn \coloneqq \E(x)$, $\mu_\out \coloneqq \E(y)$, $\nu_\inn \coloneqq \Var(x)$, and $\nu_\out \coloneqq \Var(y)$.  The notation \texttt{Conv\{1D,2D,3D\}} is used to refer to \texttt{Conv1D}, \texttt{Conv2D}, and \texttt{Conv3D}, and analogously for other layer types.  Inputs to each layer are assumed to be independent and normally distributed.  Although these assumptions may not always hold exactly, experiments show that AutoInit models signal propagation across different types of networks well in practice.

\paragraph{Convolution and Dense Layers}
\label{sec:conv_dense}
The analysis in the next paragraph applies to \texttt{Conv\{1D,2D,3D\}}, \texttt{DepthwiseConv\{1D,2D\}}, and \texttt{Dense} layers, since convolution layers are dense layers with sparse connectivity.  Notation and derivation are inspired by that of \citet{glorot2010understanding} and \citet{he2015delving}.  

A feedforward layer can be written as $y = W x + b$, where $x$ is the input, $W$ is a $\texttt{fan\_out} \times \texttt{fan\_in}$ weight matrix, $b$ is a vector of biases, and $y$ is the result.  Assume the elements of $W$ are mutually independent and from the same distribution, and likewise for the elements of $x$.  Further assume that $W$ and $x$ are independent of each other. The outgoing mean can then be written as $\mu_\out = \E(W)\mu_\inn$. For the outgoing variance, letting $W$ have zero mean and expanding the product of independent random variables yields $\nu_\out = \texttt{fan\_in}\Var(W)(\nu_\inn + \mu_\inn^2)$.  Sampling the weights $W$ according to
\begin{equation} \label{eq:variance_scaling}
    W \sim \mathcal{N}\left(0, \frac{1}{\sqrt{\texttt{fan\_in}(\nu_\inn + \mu_\inn^2)}}\right)
\end{equation}
or
\begin{equation} \label{eq:variance_scaling_1}
    {\small
    W \sim \mathcal{U}\left(-\frac{\sqrt{3}}{\sqrt{\texttt{fan\_in}(\nu_\inn + \mu_\inn^2)}}, \frac{\sqrt{3}}{\sqrt{\texttt{fan\_in}(\nu_\inn + \mu_\inn^2)}}\right)}
\end{equation}
is sufficient to ensure that 
\begin{equation}
    \mu_\out = 0 \textrm{ and } \nu_\out = 1.
\end{equation}


\paragraph{Activation Functions}
\label{sec:afn}
The analysis in the next paragraph accounts for all activation functions in TensorFlow, including \texttt{elu}, \texttt{exponential}, \texttt{gelu}, \texttt{hard\_sigmoid}, \texttt{LeakyReLU}, \texttt{linear}, \texttt{PReLU}, \texttt{ReLU}, \texttt{selu}, \texttt{sigmoid}, \texttt{softplus}, \texttt{softsign}, \texttt{swish}, \texttt{tanh}, and \texttt{ThresholdedReLU} \citep[by][respectively]{elu, hendrycks2016gaussian, maas2013rectifier, he2015delving, nair2010rectified, selu, DBLP:conf/iclr/RamachandranZL18, elfwing2018sigmoid, courbariaux2015binaryconnect}, and in fact extends to any integrable \texttt{Activation} function $f$.

Let $p_\mathcal{N}(x; \mu, \sigma)$ denote the probability density function of a Gaussian distribution with mean $\mu$ and standard deviation $\sigma$.  By the law of the unconscious statistician,
\begin{align}
    \mu_\out &= \int_{-\infty}^\infty f(x) p_\mathcal{N}(x; \mu_\inn, \sqrt{\nu_\inn}) \diff x, \\
    \nu_\out &= \int_{-\infty}^\infty f(x)^2 p_\mathcal{N}(x; \mu_\inn, \sqrt{\nu_\inn}) \diff x - \mu_\out^2.
\end{align}
These integrals are computed for an arbitrary activation function $f$ with adaptive quadrature, a well-established numerical integration approach that approximates integrals using adaptively refined subintervals \cite{piessens2012quadpack, 2020SciPy-NMeth, numpy-harris2020array}.

\paragraph{Dropout Layers}
\label{sec:dropout}
Dropout layers randomly set \texttt{rate} percentage of their inputs to zero \cite{srivastava2014dropout}.  Therefore, 
\begin{equation}
    \mu_\out = \mu_\inn(1 - \texttt{rate}) \textrm{ and } \nu_\out = \nu_\inn(1 - \texttt{rate}).
\end{equation}
However, this analysis only applies to \texttt{SpatialDropout\{1D,2D,3D\}} layers.  For regular \texttt{Dropout} layers, TensorFlow automatically scales the values by $1 / (1 - \texttt{rate})$ to avoid a mean shift towards zero.\footnote{\url{https://github.com/tensorflow/tensorflow/blob/v2.5.0/tensorflow/python/keras/layers/core.py\#L149-L150}}  Adjusting for this change gives 
\begin{equation}
    \mu_\out = \mu_\inn \textrm{ and } \nu_\out = \nu_\inn/(1 - \texttt{rate}).
\end{equation}

\paragraph{Pooling Layers}
\label{sec:pooling}
The same approach applies to all commonly used pooling layers, including \texttt{AveragePooling\{1D,2D,3D\}}, \texttt{MaxPooling\{1D,\allowbreak2D,3D\}}, \texttt{GlobalAveragePooling\{1D,2D,3D\}}, and \texttt{GlobalMaxPooling\{1D,2D,3D\}}.

Let $op(\cdot)$ be the average operation for an average pooling layer, and the maximum operation for a max pooling layer.  Define $K$ to be the pool size of the layer.  For standard 1D, 2D, and 3D pooling layers, $K$ would equal $k$, $k \times k$, and $k \times k \times k$, respectively.  The global pooling layers can be seen as special cases of the standard pooling layers where the pool size is the same size as the input tensor, except along the batch and channel dimensions.  Analytically, the outgoing mean and variance can be expressed as
\begin{align}
    \begin{split}
        \mu_\out &= \idotsint_{\mathbb{R}^{K}} op(x_1, x_2, \ldots, x_{K}) \cdot ~   \\
        &\qquad \prod_{i=1}^{K} p_\mathcal{N}(x_i; \mu_\inn, \sqrt{\nu_\inn}) \diff x_1 \diff x_2 \cdots \diff x_{K},
    \end{split}\\
    \begin{split}
        \nu_\out &= \idotsint_{\mathbb{R}^{K}} op(x_1, x_2, \ldots, x_{K})^2 \cdot ~  \\ 
        &\qquad \prod_{i=1}^{K} p_\mathcal{N}(x_i; \mu_\inn, \sqrt{\nu_\inn}) \diff x_1 \diff x_2 \cdots \diff x_{K} - \mu_\out^2,
    \end{split}
\end{align}
where the $x_i$ represent tensor entries within a pooling window.  Unfortunately, even a modest $3 \times 3$ pooling layer requires computing nine nested integrals, which is prohibitively expensive.  In this case, a Monte Carlo simulation is appropriate.
Sample $x_{1_j}, x_{2_j}, \ldots x_{K_j}$ from $\mathcal{N}(\mu_\inn, \sqrt{\nu_\inn})$ for $j = 1, \ldots, S$ and return
\begin{align}
    \mu_\out &= \frac{1}{S} \sum_{j=1}^S op(x_{1_j}, x_{2_j}, \ldots, x_{K_j}),\\
    \nu_\out &= \frac{1}{S} \sum_{j=1}^S op(x_{1_j}, x_{2_j}, \ldots, x_{K_j})^2 - \mu_\out.
\end{align}

\paragraph{Normalization Layers}
\label{sec:normalization}
\texttt{BatchNormalization}, \texttt{LayerNormalization}, and \texttt{GroupNormalization} normalize the input to have mean zero and variance one \cite{ioffe2015batch, ba2016layer, wu2018group}.  Thus, 
\begin{equation}
    \mu_\out = 0 \textrm{ and } \nu_\out = 1.
\end{equation}

\paragraph{Arithmetic Operators}
\label{sec:arithmetic}
Assume the input tensors $x_1, x_2, \ldots, x_N$ with means $\mu_{\inn_1}, \mu_{\inn_2}, \ldots, \mu_{\inn_N}$ and variances $\nu_{\inn_1}, \nu_{\inn_2}, \ldots, \nu_{\inn_N}$ are independent.  The following mean and variance mapping functions are derived.
    For an \texttt{Add} layer,
    \begin{equation}\mu_\out = \sum_{i=1}^N \mu_{\inn_i} \textrm{ and } \nu_\out = \sum_{i=1}^N \nu_{\inn_i}.\end{equation}
    For an \texttt{Average} layer,
    \begin{equation}\mu_\out = \frac{1}{N}\sum_{i=1}^N \mu_{\inn_i} \textrm{ and } \nu_\out = \frac{1}{N^2}\sum_{i=1}^N
    \nu_{\inn_i}.\end{equation}
    For a \texttt{Subtract} layer,
    \begin{equation}\mu_\out = \mu_{\inn_1} - \mu_{\inn_2} \textrm{ and } \nu_\out = \nu_{\inn_1} + \nu_{\inn_2}.\end{equation}
    Finally, for a \texttt{Multiply} layer,
    \begin{equation}\mu_\out = \prod_{i=1}^N \mu_{\inn_i} \textrm{ and } \nu_\out = \prod_{i=1}^N (\nu_{\inn_i} + \mu_{\inn_i}^2) - \prod_{i=1}^N \mu_{\inn_i}^2.\end{equation}


\paragraph{Concatenation Layers}
\label{sec:concatenation}
Assume the inputs $x_1, x_2, \ldots, x_N$ with means $\mu_{\inn_1}, \mu_{\inn_2}, \ldots, \mu_{\inn_N}$ and variances $\nu_{\inn_1}, \nu_{\inn_2}, \ldots, \nu_{\inn_N}$ are independent, and let input $x_i$ have $C_i$ elements.  Then for a \texttt{Concatenate} layer,
\begin{align}
    \mu_\out &= \frac{1}{\sum C_i} \sum_{i=1}^N C_i \mu_{\inn_i},\\
    \nu_\out &= \frac{1}{\sum C_i}\sum_{i=1}^N C_i(\nu_{\inn_i} + \mu_{\inn_i}^2) - \mu_\out^2.
\end{align}

\paragraph{Recurrent Layers}
\label{sec:recurrent}
A Monte Carlo simulation can be used to estimate the outgoing mean and variance for recurrent layers, including \texttt{GRU}, \texttt{LSTM}, and \texttt{SimpleRNN} \cite{hochreiter1997long, chung2014empirical}.  Recurrent layers often make use of activation functions like sigmoid and tanh that constrain the scale of the hidden states.  Because of this practice, recurrent layers should be initialized with a default scheme or according to recent research in recurrent initialization \cite{chen2018dynamical, gilboa2019dynamics}.  AutoInit will then estimate the outgoing mean and variance in order to inform appropriate weight scaling elsewhere in the network.

\paragraph{Padding Layers}
\label{sec:zero_padding}
\texttt{ZeroPadding\{1D,2D,3D\}} layers augment the borders of the input tensor with zeros, increasing its size.  Let $z$ be the proportion of elements in the tensor that are padded zeros.  Then $z = (\texttt{padded\_size} - \texttt{original\_size}) / \texttt{padded\_size}$, and
\begin{equation}
    \mu_\out = \mu_\inn(1-z) \textrm{ and } \nu_\out = \nu_\inn(1-z).
\end{equation}

\paragraph{Shape Adjustment Layers}
\label{sec:shape_adjustment}
Many layers alter the size or shape of the input tensor but do not change the distribution of the data.  These layers include \texttt{Flatten}, \texttt{Permute}, \texttt{Reshape}, \texttt{UpSampling\{1D,2D,3D\}}, and \texttt{Cropping\{1D,2D,3D\}} layers.  The same is true of TensorFlow API calls \texttt{tf.reshape}, \texttt{tf.split}, and \texttt{tf.transpose}.  For these layers, 
\begin{equation}
    \mu_\out = \mu_\inn \textrm{ and } \nu_\out = \nu_\inn.
\end{equation}

\paragraph{Input Layers}
\label{sec:input}
An \texttt{InputLayer} simply exposes the model to the data, therefore 
\begin{equation}
    \mu_\out = \mu_{\mathrm{data}} \textrm{ and } \nu_\out = \nu_{\mathrm{data}}.
\end{equation} 
TensorFlow allows nesting models within other models.  In this use case where the \texttt{InputLayer} does not directly connect to the training data, 
\begin{equation}
    \mu_\out = \mu_\inn \textrm{ and } \nu_\out = \nu_\inn.
\end{equation}

\paragraph{Matrix Multiplication}
A call to \texttt{tf.matmul} takes input tensors $x_1$ and $x_2$ of shape $\cdots \times m \times n$ and $\cdots \times n \times p$ and produces the output tensor $x_{\out}$ of shape $\cdots \times m \times p$ with entries
\begin{equation}
    x_{\out_{\cdots,i,j}} = \sum_{k = 1}^n x_{1_{\cdots,i,k}} x_{2_{\cdots,k,j}}.
\end{equation}
Assuming independent matrix entries, the output statistics can then be calculated as
\begin{align}
    \mu_\out &= n \mu_{\inn_1} \mu_{\inn_2},\\
    \nu_\out &= n \left((\nu_{\inn_1} + \mu_{\inn_1}^2)(\nu_{\inn_2} + \mu_{\inn_2}^2) - \mu_{\inn_1}^2\mu_{\inn_2}^2\right).
\end{align}

\paragraph{Reduction Operators}
A call to $\texttt{tf.reduce\_mean}$ reduces the size of the input tensor by averaging values along one or more axes.  For example, averaging an input tensor of shape $128 \times 8 \times 8 \times 256$ along axes 1 and 2 would produce an output tensor of shape $128 \times 1 \times 1 \times 256$.  Let $D$ represent the product of the length of the axes being averaged over (in the example above, $8 \times 8 = 64$).  The output tensor has
\begin{equation}
    \mu_\out = \mu_\inn \textrm{ and } \nu_\out = \nu_\inn / D.
\end{equation}
The function $\texttt{tf.reduce\_sum}$ performs similarly, summing entries instead of averaging them.  In this case,
\begin{equation}
    \mu_\out = D\mu_\inn \textrm{ and } \nu_\out = D\nu_\inn.
\end{equation}

\paragraph{Maintaining Variance $\mathbf{\neq 1}$.}
In Algorithm \ref{alg:autoinit}, AutoInit initializes weights so that the output signal at each layer has mean zero and variance one.  Although signal variance $\nu = 1$ is sufficient for stable signal propagation, it is not a necessary condition.  Indeed, other values for the signal variance $\nu$ could be utilized, as long as $\nu$ remains consistent throughout the network.  If a different $\nu$ is desired, weights can be initialized according to Equation \ref{eq:variance_scaling} or \ref{eq:variance_scaling_1} and then multiplied by $\sqrt{\nu}$. For instance, such a modification was done for the CoAtNet model in Section~\ref{sec:coatnet}, resulting in slightly improved final performance.

\section{Convolutional, Residual, and Transformer Network Experiment Details}
\label{sec:training_details}

This appendix contains implementation details for the experiments in Sections \ref{sec:convolutional}-\ref{sec:autoinit_vs_lsuv}.

\paragraph{All-CNN-C} The training setup follows that of \citet{springenberg2015striving} as closely as possible.  The network is trained with SGD and momentum 0.9.  The dropout rate is 0.5 and weight decay as L2 regularization is 0.001.  The data augmentation involves featurewise centering and normalizing, random horizontal flips, and random $32 \times 32$ crops of images padded with five pixels on all sides.  The initial learning rate is 0.01 and is decreased by a factor of 0.1 after epochs 200, 250, and 300 until training ends at epoch 350.

Because \citet{springenberg2015striving} did not specify how they initialized their weights, the networks are initialized with the ``Glorot Uniform'' strategy \citep[also called Xavier initialization; ][]{glorot2010understanding}, where weights are sampled from $\mathcal{U}\left(-\frac{\sqrt{6}}{\sqrt{\texttt{fan\_in} + \texttt{fan\_out}}}, \frac{\sqrt{6}}{\sqrt{\texttt{fan\_in} + \texttt{fan\_out}}}\right)$. This initialization is the default setting in TensorFlow,\footnote{\url{https://github.com/tensorflow/tensorflow/blob/v2.5.0/tensorflow/python/keras/layers/convolutional.py\#L608-L609}} and is sufficient to replicate the results reported by \citet{springenberg2015striving}. 

\paragraph{Residual Networks}
The networks are optimized with SGD and momentum 0.9.  Dropout is not used, and weight decay is 0.0001.  Data augmentation includes a random horizontal flip and random $32 \times 32$ crops of images padded with four pixels on all sides.

\paragraph{CoAtNet}

A smaller variant of the CoAtNet architecture\footnote{\url{https://github.com/leondgarse/keras_cv_attention_models/blob/v1.3.0/keras_cv_attention_models/coatnet/coatnet.py#L199}} was used in order to fit the model and data on the available GPU memory.  The architecture has three convolutional blocks with 64 channels, four convolutional blocks with 128 channels, six transformer blocks with 256 channels, and three transformer blocks with 512 channels.  This architecture is slightly deeper but thinner than the original CoAtNet-0 architecture, which has two convolutional blocks with 96 channels, three convolutional blocks with 192 channels, five transformer blocks with 384 channels, and two transformer blocks with 768 channels \cite{dai2021coatnet}.  The models are otherwise identical.

The training hyperparameters were inspired by \citet{wightman2021resnet} and are common in the literature.  Specifically, images were resized to $160 \times 160$.  The learning rate schedule was increased linearly from $1e^{-4}$ to $4e^{-4}$ for six epochs and it then followed a cosine decay until epoch 105.  Weight decay was set to 0.02 times the current learning rate at each epoch.  The model was trained with batch size 256 and optimized with AdamW \cite{loshchilov2017decoupled_adamw}.  Data augmentation included RandAugment applied twice with a magnitude of six \cite{cubuk2020randaugment}.  Mixup and Cutmix were also used with alpha 0.1 and 1.0, respectively \cite{zhang2017mixup, yun2019cutmix}.  The training images were augmented with random resized crops \cite{szegedy2015going} that were at minimum 8\% of the original image; after training the model was evaluated on 95\% center crops.

As discussed in Appendix \ref{sec:mean_variance_estimation}, AutoInit maintains signal variance $\nu=1$, but it is also possible to adjust $\nu$ if desired.  In the CoAtNet experiments, $\nu = 0.01$ was found to give the best performance among $\nu = \{1, 0.1, 0.01, 0.001\}$.  The experiment removing normalization layers used the default of $\nu = 1$.

\paragraph{ImageNet} The experiment in Section \ref{sec:imagenet} used the same training setup as the experiments with CoAtNet on Imagenette in the previous paragraph except for two changes.  The batch size was 2{,}048 (512 per GPU across four GPUs), and the maximum learning rate was $3.2e^{-2}$.

\paragraph{Data-Dependent Initialization Comparison}
In the experiments in Section \ref{sec:autoinit_vs_lsuv}, a learning rate schedule inspired by superconvergence \cite{smith2019super} was used to save time.  The learning rate increases linearly to 0.1 during the first five epochs, and then decreases linearly for 20 epochs, after which training ends.  The weight decay for All-CNN-C was also decreased by a factor of 10.  This modification is common when networks are trained with superconvergence \cite{smith2019super}.

\paragraph{Initialization Time}
It is important to note that AutoInit does not incur a significant overhead.  Each layer must be visited once to be initialized, so the complexity is $O(L)$ where $L$ is the number of layers.  For example: All-CNN-C, ResNet-56, and ResNet-164 take 1, 33, and 106 seconds to initialize.  The costs are hardware-dependent, but only spent once, and are small compared to the cost of training.

\section{Neural Architecture Search Experiment Details}
\label{sec:enn_details}

\begin{figure}
    \centering
    \includegraphics[width=0.9\linewidth]{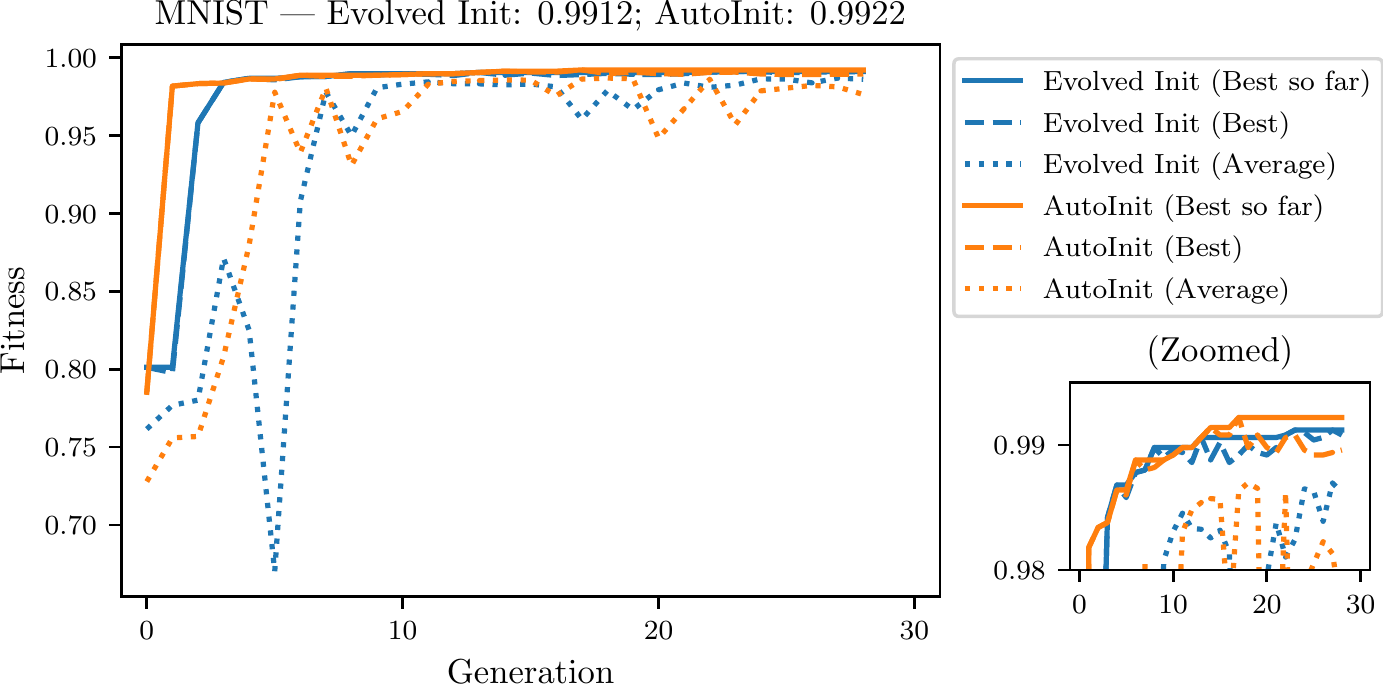}
    \includegraphics[width=0.9\linewidth]{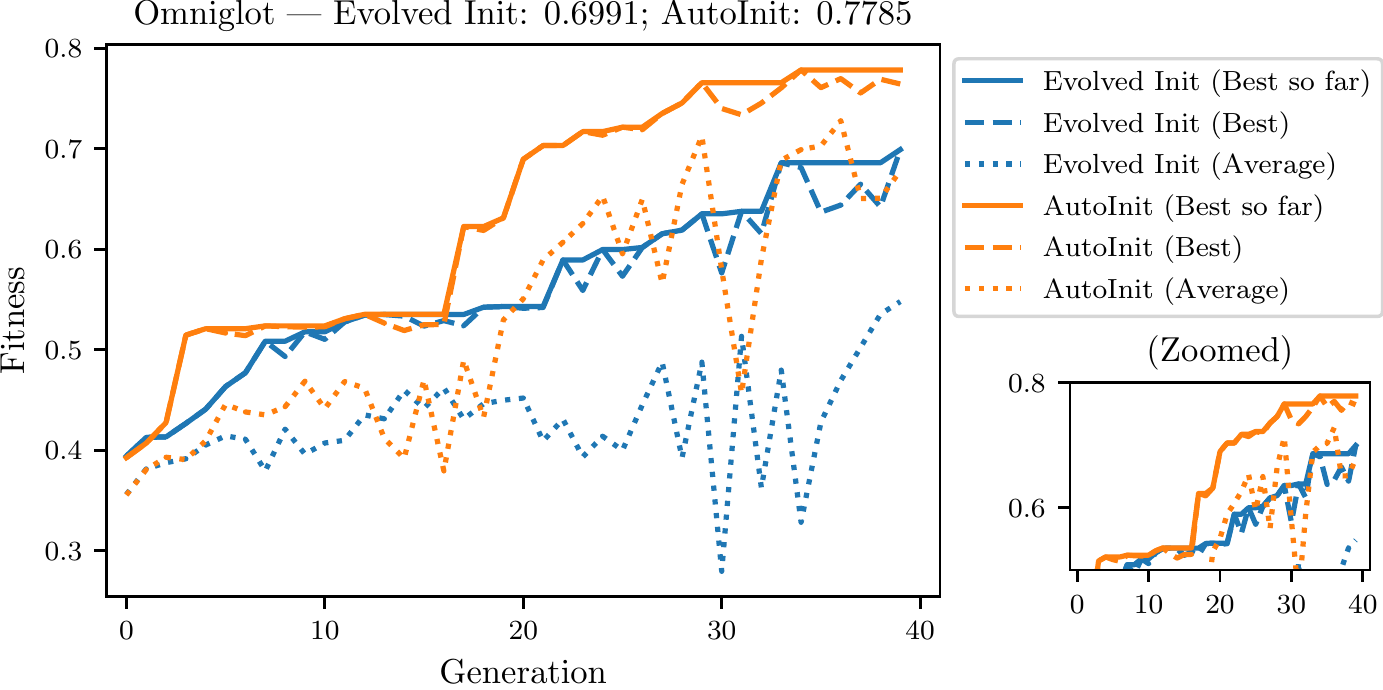}
    \includegraphics[width=0.9\linewidth]{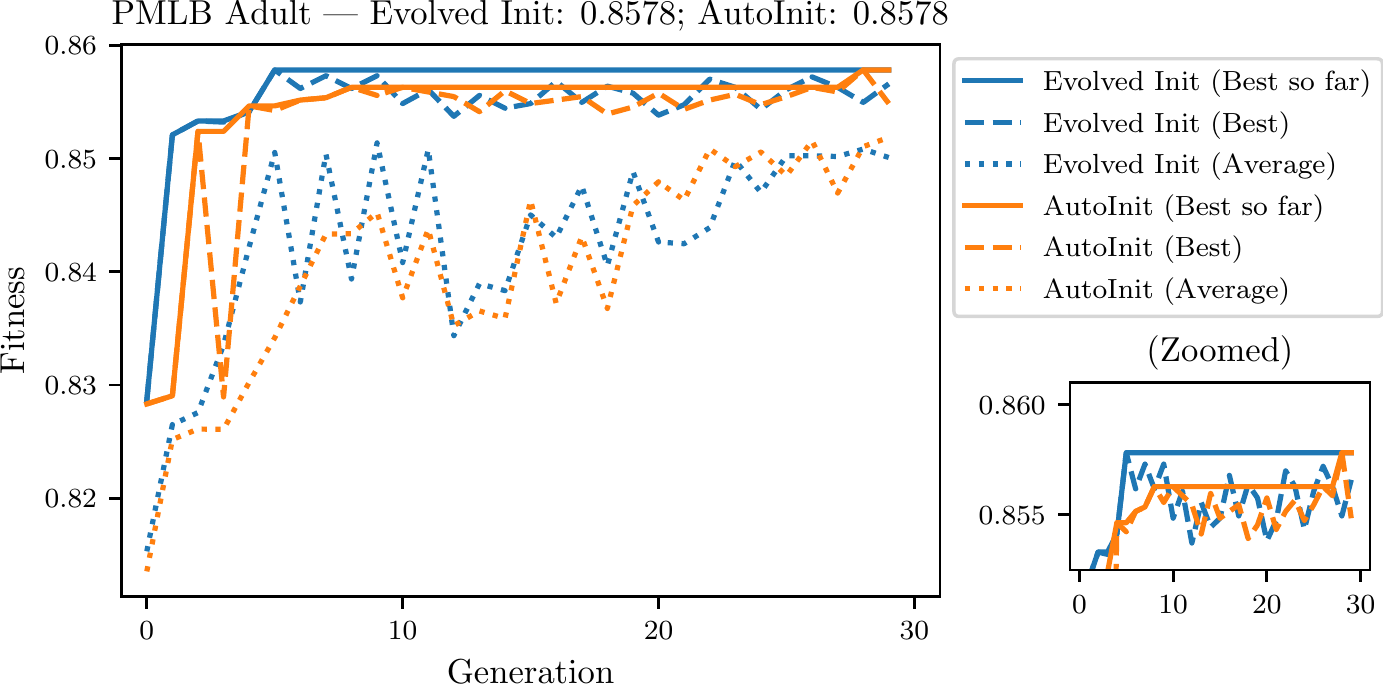}
    \includegraphics[width=0.9\linewidth]{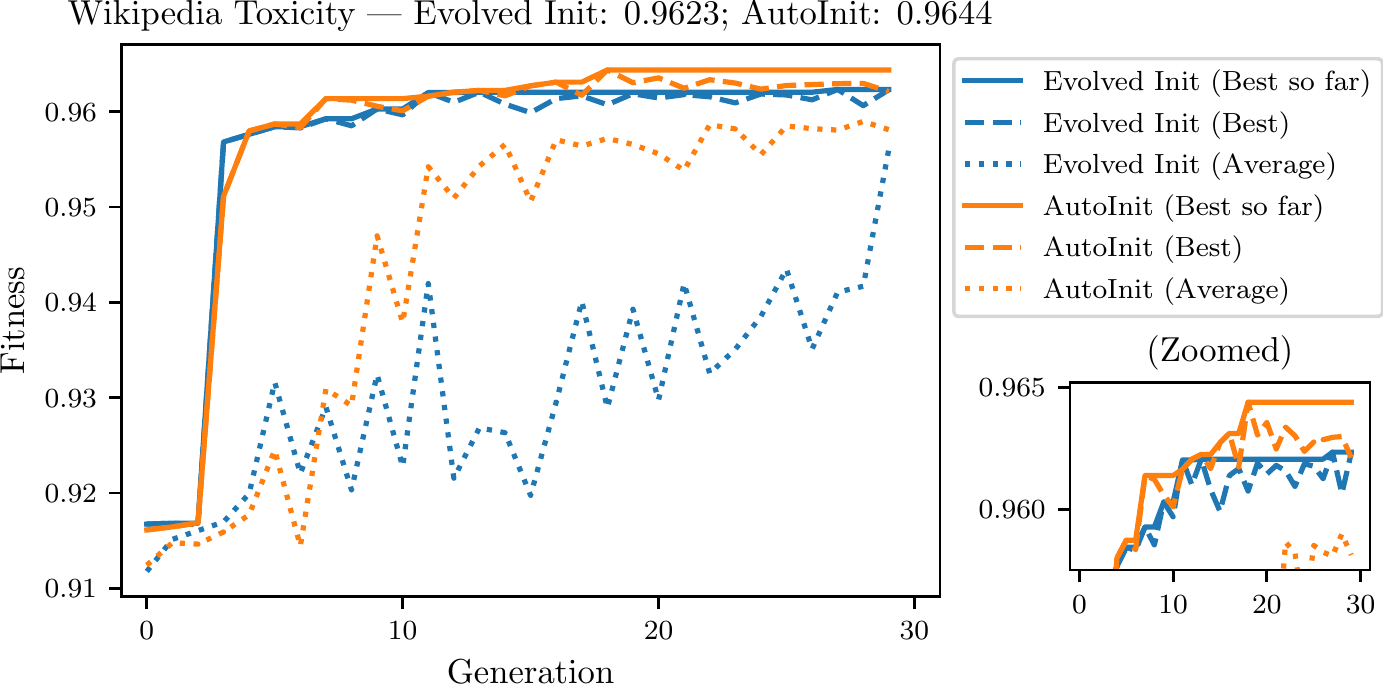}
    \includegraphics[width=0.9\linewidth]{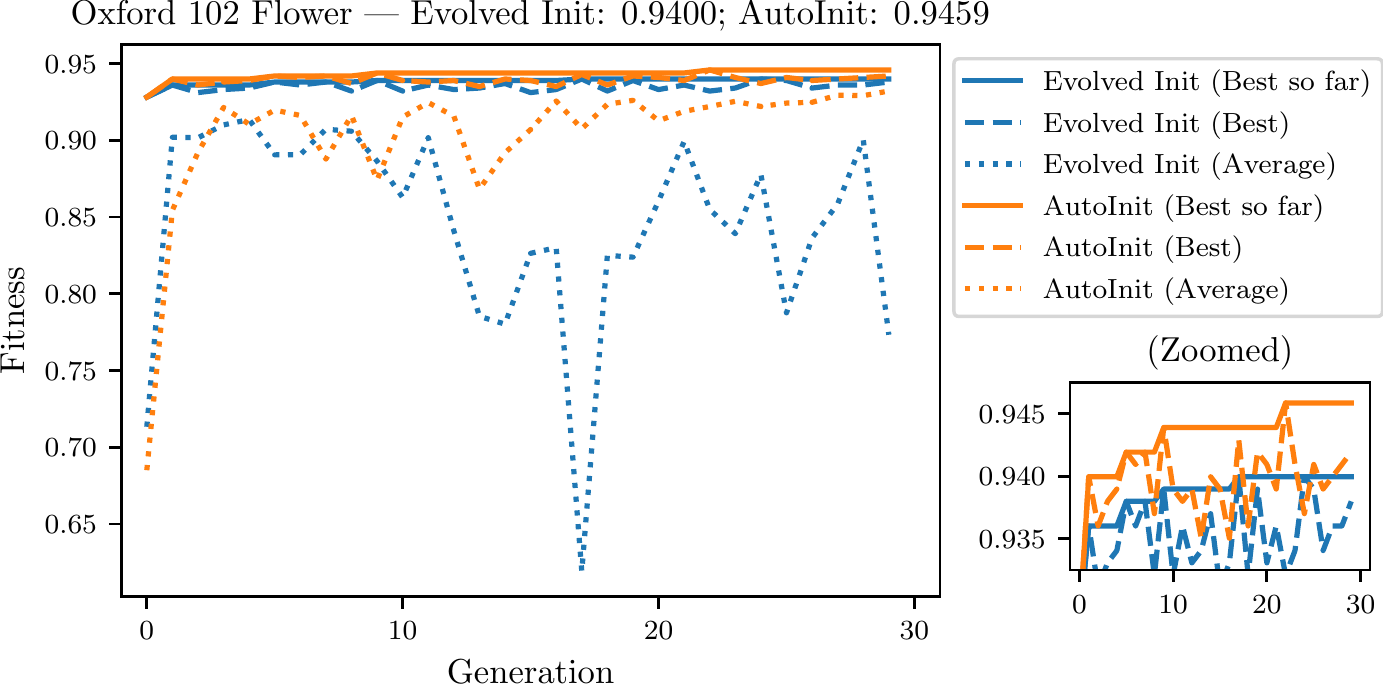}
    \caption{Progress of neural architecture search in the five tasks.  The data is the same as that in Figure~\ref{fig:enn_results}, but this plot also shows how AutoInit can stabilize mean population performance, leading to more reliable discovery of powerful models.}
    \label{fig:enn_results_detail}
\end{figure}

This appendix contains implementation details for the experiments in Section \ref{sec:nas}.  The progress of CoDeepNEAT in these experiments was shown in Figure \ref{fig:enn_results} in the main text.  Figure \ref{fig:enn_results_detail} shows this same data but in greater detail.  Table \ref{tab:hyperparameters} contains the training hyperparameters, neural network layers, evolutionary hyperparameters, and mutation probabilities used in each of the five tasks.  The five tasks used with CoDeepNEAT are:

\nocite{kingma2014adam}

\begin{table*}
    \centering
    \begin{adjustbox}{max width=\linewidth}
    \begin{tabular}{llllll}
        \toprule
         & \textbf{MNIST} & \textbf{Omniglot} & \textbf{PMLB Adult} & \textbf{Wikipedia Toxicity} & \textbf{Oxford 102 Flower} \\ \midrule 
        \multicolumn{6}{l}{\textbf{Training Hyperparameters}} \\
        Activation & \multicolumn{5}{c}{All domains: \{ReLU, Linear, ELU, SELU\}} \\
        Batch Size & 128 & 1000 iterations & 32 & 128 & 8 \\
        Dropout Rate & $[0.0, 0.7]$ & $[0.0, 0.7]$ & $[0.0, 0.9]$ & $[0.0, 0.5]$ & $[0.0, 0.9]$ \\
        Epochs & 5 & 3 & 25 & 3 & 30 \\
        Filters/Units & [16, 96] & [16, 96] & [64, 192] & [64, 192] & [64, 192] \\
        Kernel Reg. & L2: $[10^{-9}, 10^{-3}]$ & L2: $[10^{-9}, 10^{-3}]$ & \{L1, L2\}: $[10^{-9}, 10^{-3}]$ & \{L1, L2\}: $[10^{-9}, 10^{-3}]$ & \{L1, L2\}: $[10^{-7}, 10^{-3}]$ \\
        Kernel Size & \{1, 3\} & \{1, 3\} & N/A & \{1, 3, 5, 7\} & N/A \\
        Learning Rate & $[10^{-4}, 10^{-2}]$ & $[10^{-4}, 10^{-3}]$ & $[10^{-4}, 10^{-2}]$ & $[10^{-4}, 10^{-2}]$ & $[10^{-4}, 10^{-1}]$ \\
        Optimizer & Adam & Adam & Adam & Adam & SGD (Nesterov=0.9) \\
        Weight Init. & \multicolumn{5}{c}{All domains: \{Glorot Normal, Glorot Uniform, He Normal, He Uniform\} or \textbf{AutoInit}} \\ \midrule
        
        \multicolumn{6}{l}{\textbf{Neural Network Layers}} \\
        Add & \checkmark & & \checkmark & \checkmark \\
        Concatenate & & & \checkmark & \checkmark \\
        Conv1D & & & & \checkmark & \\
        Conv2D & \checkmark & \checkmark & & & \\
        Dense & & & \checkmark & & \checkmark \\
        Dropout & \checkmark & \checkmark & \checkmark & & \checkmark \\
        GRU & & & & \checkmark & \\
        LSTM & & & & \checkmark & \\
        MaxPooling1D & & & & \checkmark & \\
        MaxPooling2D & \checkmark & \checkmark \\
        SpatialDropout1D & & & & \checkmark & \\
        WeightedSum & & \checkmark \\ \midrule
        
        \multicolumn{6}{l}{\textbf{Evolutionary Hyperparameters}} \\
        Elitism (B) & 0.4 & 0.4 & 0.4 & 0.2 & 0.1 \\
        Elitism (M) & 0.4 & 0.4 & 0.4 & 0.2 & 0.4 \\ 
        Evaluations (B) & 4 & 4 & 4 & 4 & 1 \\
        Generations & 30 & 40 & 30 & 30 & 30 \\
        Population Size (B) & 22 & 22 & 22 & 22 & 20 \\
        Population Size (M) & 56 & 56 & 56 & 56 & 20 \\
        Preserved Networks & 12 & 12 & 12 & 12 & 1 \\ 
        Species (B) & 1 & 1 & 1 & 1 & 1 \\
        Species (M) & 4 & 4 & 4 & 4 & 2 \\ \midrule
        
        \multicolumn{6}{l}{\textbf{Mutation Probabilities}} \\
        Change Hyperparam. & 0.25 & 0.25 & 0.25 & 0.5 & 0.5 \\
        New Connection (B) & 0.12 & 0.12 & 0.12 & 0.2 & 0.12 \\
        New Connection (M) & 0.08 & 0.08 & 0.08 & 0.2 & 0.08 \\
        New Layer (M) & 0.08 & 0.08 & 0.08 & 0.2 & 0.08 \\
        New Node (B) & 0.16 & 0.16 & 0.16 & 0.2 & 0.16 \\

        \bottomrule
    \end{tabular}
    \end{adjustbox}
    \caption{Configuration of neural architecture search experiments in the five tasks.  Entries with a (B) or (M) suffix apply to CoDeepNEAT blueprints or modules, respectively.  These values were found to work well in preliminary experiments.  When AutoInit is applied to an evolved network, it replaces the weight initialization method selected by evolution, but the setup otherwise remains unchanged.  The neural architecture search experiments were designed to show that AutoInit improves performance in a wide variety of settings, including those with different data modalities, network topologies, computational complexities, and hyperparameter configurations.}
    \label{tab:hyperparameters}
\end{table*}

\paragraph{Vision} The MNIST dataset contains 28x28 grayscale images of handwritten digits 0-9.  There are 60,000 training images (5,000 of which were used for validation) and 10,000 test images \cite{lecun2010mnist}.  MNIST is used under the Creative Commons Attribution-Share Alike 3.0 license.

\paragraph{Language} In the Wikipedia Toxicity dataset, the task is to classify English Wikipedia comments as toxic or healthy contributions \cite{wulczyn2017ex}.  The dataset contains 92,835, 31,227, and 30,953 comments in the training, validation, and test sets, respectively.  

\paragraph{Tabular} In the Adult dataset \cite{kohavi1996scaling} from the Penn Machine Learning Benchmarks repository \citep[PMLB;][]{Olson2017PMLB}) the task is to predict whether an individual makes over \$50K per year based on 14 features.  Out of 48,842 total instances, 20\% are randomly separated to create a test set.  The dataset is used under the MIT License.

\paragraph{Multi-Task} The Omniglot dataset contains handwritten characters in 50 different alphabets \cite{lake2015human}; classifying characters in each alphabet is a natural multi-task problem.  The characters are $105 \times 105$ grayscale images, and there are 20 instances of each character.  To save compute resources, 20 of the 50 alphabets were randomly selected for experiments.  A fixed training, validation, and testing split of 50\%, 20\%, and 30\% was used with each task.  The learning rate decays as $\texttt{learning\_rate} = 0.1^{\texttt{epoch}/10}*\texttt{initial\_learning\_rate}$ during training.  The dataset is used under the MIT License.

\paragraph{Transfer Learning} A DenseNet-121 network is first pretrained on the ImageNet dataset \cite{deng2009imagenet, huang2017densely}. Models are then evolved to utilize its embeddings to classify images in the Oxford 102 Flower dataset, consisting of 102 types of flowers found in the United Kingdom \cite{Nilsback08}.  Each class has between 40 and 258 images; the training and validation sets have 10 images per class, and the test set contains the remaining images from the dataset.  During training, the weight decay (L2 loss) is scaled by the current learning rate.  Images are also augmented to improve generalization performance.  Images are randomly flipped horizontally, rotated up to 40 degrees, shifted up/down and left/right up to 20\%, and shear intensity and zoom range varied up to 20\%.

\section{Statistical Significance of Results in Activation Function Meta-Learning}

\label{sec:stat_sig}

Sampling activation functions from the PANGAEA search space results in a distribution of possible models for each weight initialization strategy.  Comparing the empirical distribution functions (EDFs) induced by each initialization strategy makes it possible to quantify the importance of the initialization \cite{radosavovic2019network}.

\begin{figure}
    \centering
    \includegraphics[width=\linewidth]{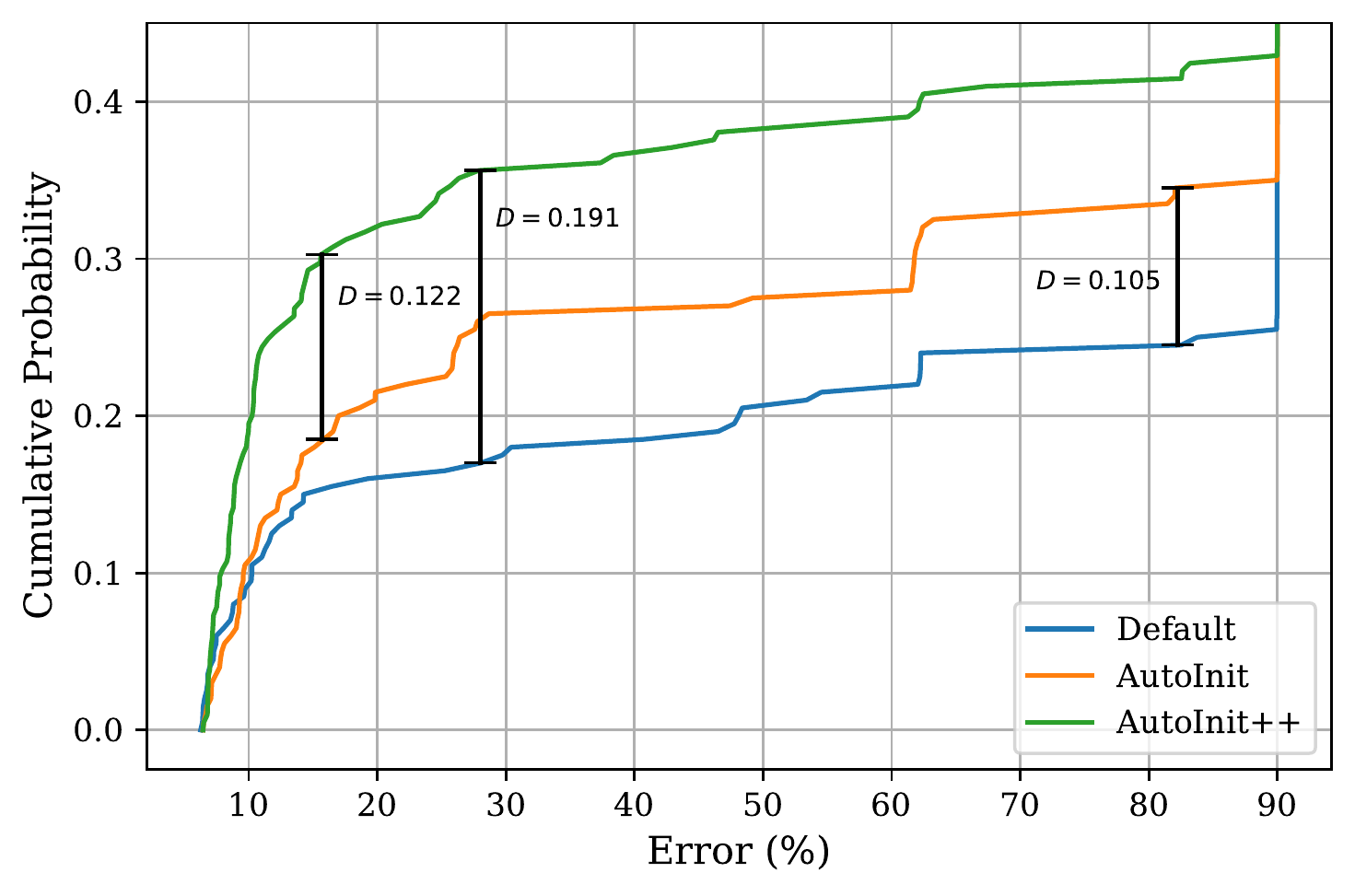}
    \caption{Error EDFs for PANGAEA activation functions when using different weight initialization strategies.  The Kolmogorov-Smirnov statistic $D$ quantifies the maximum vertical distance between the EDFs, and shows that proper initialization provides a measurable increase in expected performance.  Notice that the $x$-axis shows percent error, and not accuracy as in Figure~\ref{fig:random_afns_hist}.}
    \label{fig:afn_cdf}
\end{figure}

Given $n$ activation functions with errors $\{e_i\}$, the EDF $F(e) = \frac{1}{n}\sum_{i=1}^n\mathbf{1}[e_i < e]$ gives the fraction of activation functions that result in error less than $e$.  Let $F_\textrm{default}$, $F_\textrm{AutoInit}$, and $F_\textrm{AutoInit++}$ be the EDFs for the three initialization strategies.  Figure \ref{fig:afn_cdf} plots these EDFs along with the Kolmogorov-Smirnov test statistic $D = \sup_e |F_1(e) - F_2(e)|$, which measures the maximum vertical discrepancy between two EDFs \cite{massey1951kolmogorov}.  This statistic shows that (1) AutoInit outperforms the default initialization $(D = 0.105)$; (2) AutoInit++ delivers an even greater boost in performance over the default initialization $(D = 0.191)$; and (3) AutoInit++ is measurably better than AutoInit $(D = 0.122)$, confirming that having zero Gaussian mean is a useful property for activation functions to have.

Other ways of measuring statistical significance lead to similar conclusions. For instance, consider the null hypothesis that $F_\textrm{default} = F_\textrm{AutoInit}$.  In other words, this null hypothesis states that AutoInit provides no benefit and that the accuracies obtained come from the same underlying distribution.  With the Epps-Singleton test \cite{epps1986omnibus} this null hypothesis is rejected with $p < 0.05$.  Similarly, the test rejects the null hypothesis that $F_\textrm{default} = F_\textrm{AutoInit++}$ with $p < 0.001$.  Even stronger statements can be made in the case of AutoInit++.  With the Mann-Whitney U test \cite{mann1947test}, the null hypothesis that $F_\textrm{default}(e) \geq F_\textrm{AutoInit++}(e)$ for some $e$ is rejected $(p < 0.01)$ in favor of the alternative that $F_\textrm{default}(e) < F_\textrm{AutoInit++}(e)$ for all $e$.  Similarly, the null hypothesis that $F_\textrm{AutoInit}(e) \geq F_\textrm{AutoInit++}(e)$ for some $e$ is rejected $(p < 0.05)$ in favor of the alternative that $F_\textrm{AutoInit}(e) < F_\textrm{AutoInit++}(e)$ for all $e$.  As discussed in the main text, this result states that the distribution of accuracies induced by AutoInit++ is stochastically larger than that from AutoInit or the default initialization.

\section{Future Technical Extensions}
\label{sec:technical_extensions}

AutoInit is based on understanding and utilizing the training dynamics of neural networks, leading to higher and more robust performance, and facilitating further advances in meta-learning. It can be improved and its scope broadened in several ways in the future, as outlined below.

\paragraph{Initial Weight Distributions}
AutoInit calculates appropriate weight scaling, but it does not impose a distribution from which weights are drawn (Equation \ref{eq:variance_scaling}).  All experiments in this paper used a truncated normal distribution.  In preliminary experiments, AutoInit also used untruncated normal, uniform, and orthogonal distributions, but no clear trends were observed.  Indeed, assuming weights are scaled appropriately, whether training is stable depends only on the architecture and not the distribution from which weights are sampled \cite{hanin2018neural}.  However, this conclusion applies only in limited theoretical settings; in other settings, orthogonal initialization was found to be beneficial \cite{saxe2013exact, hu2020provable}.  Whether there is a single distribution that is optimal in every case, or whether certain distributions are better-suited to different models, tasks, or layers, remains an open question, and a compelling direction for future research.

\paragraph{Variations of AutoInit}
Several variations of the core AutoInit algorithm can be devised that may improve its performance.  For example, AutoInit stabilizes signals by analyzing the forward pass of activations from the input to the output of the network.  It is possible to similarly model the backward pass of gradients from the output to the input.  Indeed, past weight initialization strategies have sometimes utilized signals in both directions  \cite{glorot2010understanding, arpit2019initialize}.  It would be interesting to find out whether AutoInit could similarly benefit from analyzing backward-propagating signals.

Alternative objectives beyond mean and variance stabilization could also be considered.  Two promising objectives are tuning the conditioning of the Fisher information matrix \cite{pennington2018spectrum} and achieving dynamical isometry \cite{xiao2018dynamical}.  Mean field theory and nonlinear random matrix theory \cite{pennington2019nonlinear} could potentially be used to implement these objectives into AutoInit.

\paragraph{Support for New Layer Types}
AutoInit calculates outgoing mean and variance estimates for the majority of layer types available in current deep learning frameworks (Appendix~\ref{sec:mean_variance_estimation}).  If AutoInit encounters an unknown layer, the default behavior is to assume that the mean and variance are not changed by that layer: $g_\texttt{layer}(\mu_\inn, \nu_\inn) = \mu_\inn, \nu_\inn$.  This fallback mechanism tends to work well; if there are only a few unknown layers, then the variance estimation will be incorrect only by a constant factor and training can proceed.  However, mean and variance estimation functions $g$ can be derived for new types of layers as they are developed, either analytically or empirically with Monte Carlo sampling, thus taking full advantage of AutoInit's ability to stabilize training in the future as well.

\paragraph{Tighter Integration with Deep Learning Frameworks}
Using AutoInit is simple in practice.  The AutoInit package provides a wrapper around TensorFlow models.  The wrapper automatically traverses the TensorFlow computation graph, calculates mean and variance estimations for each layer, and reinstantiates the model with the correct weight scaling.  However, this implementation can be streamlined.  The most effective approach would be to integrate AutoInit natively with deep learning frameworks like TensorFlow \cite{abadi2016tensorflow} and PyTorch \cite{paszke2019pytorch}.  Native integration would not just make AutoInit easier to use, it would also make it more accessible to general machine learning practitioners.  For example, TensorFlow provides a few initialization strategies that can be leveraged by changing the \texttt{kernel\_initializer} keyword in certain layers.  However, implementing other weight initialization strategies requires subclassing from the \texttt{Initializer} base class, which is both time-consuming and complicated, especially for non-experts.  Native integration would ensure that the benefits of smarter initialization are available immediately to the wider machine learning community.

\section{Computing Infrastructure}
\label{sec:compute_infrastructure}
Experiments in this paper were run in a distributed framework using StudioML software \cite{gorunner, StudioML} to place jobs on machines with NVIDIA GeForce GTX 1080 Ti and RTX 2080 Ti GPUs.  The CoAtNet experiments (Section \ref{sec:coatnet}) were run on an AWS g5.48xlarge instance with NVIDIA A10G GPUs.  Training CoAtNet on Imagenette required an average of 0.91 GPU hours per run.  Training CoAtNet on ImageNet once took 119.89 GPU hours.  Training on Imagenette instead of ImageNet therefore emit approximately $119.89 / 0.91 \approx 132$ times less carbon.  The AutoInit package is available at \url{https://github.com/cognizant-ai-labs/autoinit}.

\end{document}